\documentclass{article}

\usepackage{algorithm}
\usepackage{algpseudocode}
\usepackage{amsmath, amssymb, amsfonts, nccmath}
\usepackage[toc,page]{appendix}
\usepackage{authblk}
\usepackage{caption}
\usepackage{diagbox}
\usepackage{dsfont}
\usepackage{enumitem}
\usepackage[left=3cm,right=3cm,top=3cm,bottom=3cm]{geometry}
\usepackage[hidelinks]{hyperref}
\usepackage[utf8]{inputenc}
\usepackage{longtable}
\usepackage{makecell}
\usepackage{multirow}
\usepackage{pifont}
\usepackage{ragged2e}
\usepackage{soul}
\usepackage{subfigure}
\usepackage{svg}
\usepackage{tabularray}
\usepackage[normalem]{ulem}

\newcolumntype{C}[1]{>{\centering\arraybackslash}m{#1}}

\title{Novel Class Discovery: an Introduction and Key Concepts}

\author[1,2]{Colin Troisemaine}
\author[1]{Vincent Lemaire}
\author[1]{Stéphane Gosselin}
\author[2]{Alexandre Reiffers-Masson}
\author[1]{Joachim Flocon-Cholet}
\author[2]{Sandrine Vaton}

\affil[1]{Orange Labs, Lannion, France}
\affil[2]{Department of Computer Science, IMT Atlantique, Brest, France}

\date{}

\begin{document}

\maketitle


\begin{abstract}
    Novel Class Discovery (NCD) is a growing field where we are given during training a labeled set of known classes and an unlabeled set of different classes that must be discovered.
    In recent years, many methods have been proposed to address this problem, and the field has begun to mature.
    In this paper, we provide a comprehensive survey of the state-of-the-art NCD methods.
    We start by formally defining the NCD problem and introducing important notions.
    We then give an overview of the different families of approaches, organized by the way they transfer knowledge from the labeled set to the unlabeled set.
    We find that they either learn in two stages, by first extracting knowledge from the labeled data only and then applying it to the unlabeled data, or in one stage by conjointly learning on both sets.
    For each family, we describe their general principle and detail a few representative methods.
    Then, we briefly introduce some new related tasks inspired by the increasing number of NCD works.
    We also present some common tools and techniques used in NCD, such as pseudo labeling, self-supervised learning and contrastive learning.
    Finally, to help readers unfamiliar with the NCD problem differentiate it from other closely related domains, we summarize some of the closest areas of research and discuss their main differences.
\end{abstract}


\section{Introduction}

\begin{figure}[htbp]
	\centering
	\includegraphics[width=0.95\linewidth]{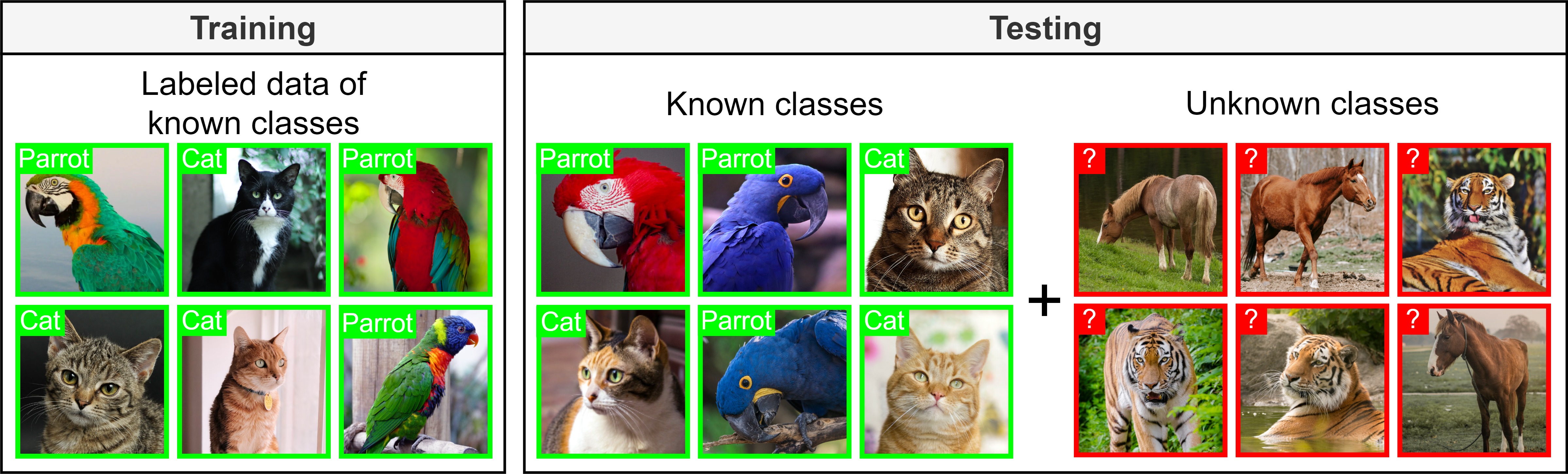}
    \caption{The open-world scenario, where new classes appear during inference.}
    \label{fig:vincent_01}
\end{figure}

In the past decade of machine learning research, many classification models have relied heavily on the availability of large amounts of labeled data for all relevant classes.
The recent success of these models is due in part to the abundance of labeled data.
However, it is not always possible to have labeled data for all classes of interest, leading researchers to consider scenarios where unlabeled data is available.
This ``open-world'' assumption is becoming increasingly more common in practical applications, where instances outside the initial set of classes may emerge \cite{oodsurvey}.
To illustrate, let's examine the scenario of Figure~\ref{fig:vincent_01}.
Here, instances from classes never seen during training appear at test time.
An ideal model should not only be able to classify the known classes (parrots and cats), but also to discover the new ones (tigers and horses).

\par \textit{What is the issue?} - In this example, a standard classification model is likely to incorrectly classify instances that fall outside the known classes as belonging to one of the known classes.
This is a well-known phenomenon of neural networks, where they can produce overconfident incorrect predictions, even in the case of semantically related inputs \cite{nguyen2015deep}.
Here, a tiger would be classified as a parrot or a cat.
For this reason, researchers are now exploring scenarios where unlabeled data is also available \cite{NodetLBCO21a, Zhou2017}.
In this survey, we will focus on one such scenario, where a labeled set of known classes and an unlabeled set of unknown classes are given during training.
The goal is to learn to categorize the unlabeled data into the appropriate classes.
This is referred to as ``Novel Class Discovery (NCD)''\footnote{In this survey, we use the term ``Novel Class Discovery'' to refer to the specific domain and not to the \textit{act of discovering novel classes}.
This name is becoming gradually more popular in the literature, but it can be confusing due to its general meaning.
It is also sometimes called ``Novel \ul{Category} Discovery''.} \cite{hsu2018learning}.
\\

\begin{figure}[htbp]
	\centering
	\includegraphics[width=0.95\linewidth]{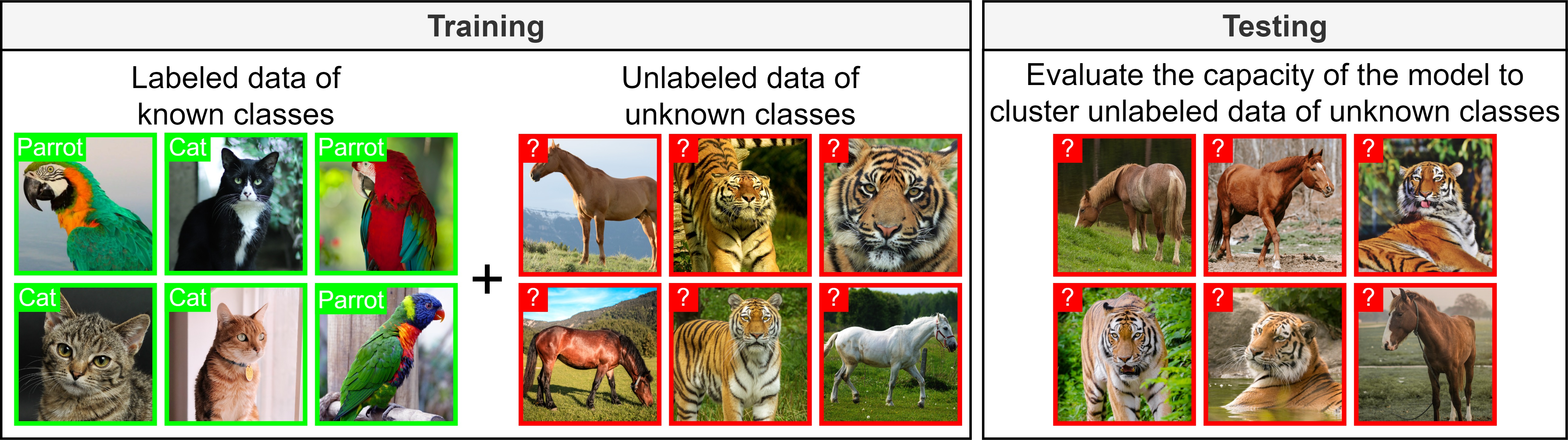}
    \caption{The Novel Class Discovery scenario, where both labeled data of known classes and unlabeled data of unknown classes are available during training.}
    \label{fig:ncd_setup}
\end{figure}

\textit{What is the usual setup of NCD?} - Illustrated in Figure~\ref{fig:ncd_setup}, the training data in NCD consists of two sets of samples: one from known classes and one from unknown classes. The test set is comprised solely of samples from unknown classes.
The NCD scenario belongs to Weakly Supervised Learning \cite{NodetLBCO21a, Zhou2017}, where methods that require all the classes to be known in advance can be distinguished from those that are able to manage classes that have never appeared during training.
As an example, in Open-World Learning (OWL) \cite{oodsurvey}, methods seek to accurately label samples of classes seen during training, while identifying samples from unknown classes.
However, the methods in OWL are generally not tasked with clustering the unknown classes and unlabeled data is left unused.
Another example is Zero-Shot Learning (ZSL) \cite{10.1145/3293318}, where the models are designed to accurately predict classes that have never appeared during training.
But some kind of description of these unknown classes is needed to be able to recognize them.
On the other hand, NCD has recently gained significant attention due to its practicality and real-world applications.

\begin{figure}[htb]
	\centering
	\includegraphics[width=0.95\linewidth]{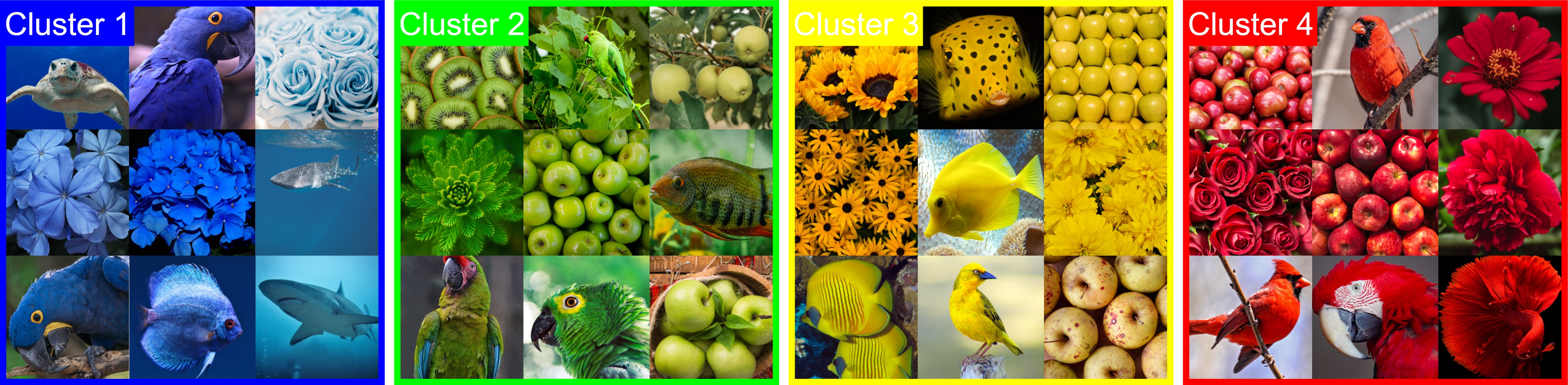}
    \caption{Example of naive solution that could be found with unsupervised clustering. The images are grouped by dominant color and not by semantic class such as bird, flower, fish, \dots}
    \label{fig:naive_clustering}
\end{figure}

\textit{Why does clustering alone fail to produce good results?} - Albeit naive, unsupervised clustering is a direct solution to the NCD problem as it can sometimes be sufficient for discovering classes in unlabeled data.
For example, many clustering methods have obtained an accuracy larger than 90\% on the MNIST dataset \cite{DMSC, depict, 1604.03628}.
But in the case of complex datasets, the literature shows that clustering fails \cite{tcl, IMC-SwAV} compared to more sophisticated approaches.
Clustering can fail for many reasons due to the assumptions that the methods make: 
spherical clusters, mixture of Gaussian distributions, shape of the data, similarity measure, etc.
Thus, the partitioning produced could be incoherent with the data or with the semantic classes; i.e. unsupervised learning is not enough in some cases.
We attempt to illustrate this idea in Figure~\ref{fig:naive_clustering}:
If the similarity measure used is highly influenced by the color of images, the clusters that are generated will likely group images based on their dominant color.
Although the clusters formed in this manner will be statistically accurate (with high similarity within the cluster and low similarity between clusters), the semantic categories will not be revealed.

As real-world datasets vary widely in nature and the desired clusters can have very different definitions, it seems impossible to create a clustering algorithm that fits all data types.
Therefore, there is a need for more refined techniques that can extract from known classes a relevant representation of a class in order to improve the clustering process.
\\

\textit{To fill these gaps} - the Novel Class Discovery domain has been proposed: it attempts to identify new classes in unlabeled data by exploiting prior knowledge from known classes.
The idea behind NCD is that by having a set of known classes, a suitable method should be able to improve its performance by extracting a general concept of what constitutes a good class.
This can, for example, take the form of a specialized similarity function or a latent space containing domain-specific features.
It is assumed that the model does not need to be able to distinguish the known from the unknown classes.
If this assumption is not made, this becomes a \textit{Generalized Category Discovery} (GCD) \cite{vaze2022generalized} problem.
Some solutions have been proposed for the NCD problem in the context of computer vision and have displayed promising results \cite{autonovel2021, han2019learning, wang2020openworld, zhong2021neighborhood}.

In most of the literature, the difficulty of a NCD problem is set by varying the number of known/unknown classes, and increasing the number of known classes is considered as a way of making the problem easier.
In \cite{li2022a}, the authors explore the influence of the semantic similarity between the classes of the labeled and unlabeled sets.
Their assumption is that if the labeled set has a high semantic similarity to the unlabeled set, the NCD problem will be easier to solve.
Intuitively, if the task is to distinguish different animal species in the unlabeled set, a set of other known animals will be beneficial, while a set of cars will not.
They prove the validity of this assumption through their experiments and find that a labeled set with low semantic similarity can even have a negative impact on the performance.
\\

\textit{Contributions and Organization of this paper} - We provide a detailed overview of Novel Class Discovery and its formulation, as well as its positioning with respect to related domains.
We outline the key components present in most NCD methods, in the form of general workflows and a study of some representative methods, organized by the way they transfer knowledge from the labeled to the unlabeled set.
Additionally, we situate related works in the context of NCD.
The remaining sections of this paper are organized as follows:
Section~\ref{sec:preliminaries} introduces relevant general knowledge and an overview of domains related to NCD.
Section~\ref{sec:taxonomy} presents a taxonomy of current NCD methods and describes some representative methods.
Section~\ref{sec:gcd} provides a brief overview of new domains derived from NCD.
Since certain techniques and tools are frequently found in NCD methods, Section~\ref{sec:tools} offers a concise description of them.
Finally, Section~\ref{sec:ncd_related_domains} highlights links and differences with related research fields before concluding.

\section{Preliminaries}
\label{sec:preliminaries}

\begin{table}[htbp]
    \centering
    \begin{tabular}{|l|p{90mm}|}
        \hline
        \textbf{Notations} & \textbf{Meaning} \\
        \hline
        $\mathcal{X}$ & the feature space in $\mathbb{R}^d$. \\
        \hline
        $X^l/X^u$ & the data samples of the labeled/unlabeled sets. \\
        \hline
        $P(X)$ & the marginal distribution of $X$. \\
        \hline
        $\mathcal{Y}^l/\mathcal{Y}^u$ & the target spaces in $\mathbb{R}^{C^l}/\mathbb{R}^{C^u}$.\\
        \hline
        $C^l/C^u$ & the number of classes in the labeled/unlabeled sets. \\
        \hline
        $Y^l/Y^u$ & the corresponding class labels of $X^l/X^u$. \\
        \hline
        $D^l/D^u$ & the labeled/unlabeled data domains, composed of a set of samples $X$ and their corresponding class labels $Y$. \\
        \hline
        $N/M$ & the number of samples in $D^l/D^u$. \\
        \hline
    \end{tabular}
    \caption{Notations frequently used in this paper and their meanings.}
    \label{tab:notations}
\end{table}

In this section, we introduce some general knowledge useful to understand most of the NCD works.
We start by briefly summarizing the history of NCD in the literature, before giving a formal definition that follows the widely used mathematical notations of \cite{zhong2021neighborhood} and \cite{han2020automatically}.
Table~\ref{tab:notations} lists some of the important notations used throughout this survey.
And we present the usual evaluation protocol and the metrics used in NCD.
\\

\textbf{A brief history of NCD:} The 2018 article of Hsu et al. \cite{hsu2018learning} can be considered the first to solve the Novel Class Discovery problem.
The authors position their work as a transfer learning task where the labels of the target set are not available and must be inferred.
Their methods, KCL \cite{hsu2018learning} and MCL \cite{hsu2019multiclass}, are still regularly used as competitors in NCD articles.
The term ``Novel Category Discovery'' was initially used by Han et al. \cite{han2020automatically} in 2020 and is another popular term to designate the NCD problem.
Building on this work, Zhong et al. defined ``Novel Class Discovery'' as a new specific setting in 2021 \cite{zhong2021neighborhood}.
\\

\textbf{A formal definition of NCD:}
During training, the data is provided in two distinct sets, a labeled set $D^l = \{ (x_i^l, y_i^l) \}_{i=1}^N$ and an unlabeled set $D^u = \{ x_i^u \}_{i=1}^M$.
Each $x_i^l \in D^l$ and $x_i^u \in D^u$ are data instances and $y_i^l \in \mathcal{Y}^l = \{ 1, \dots, C^l\ \}$ are the corresponding class labels of $D^l$.
The goal is to use both $D^l$ and $D^u$ to discover the $C^u$ novel classes, and this is usually done by partitioning $D^u$ into $C^u$ clusters and associating labels $y_i^u \in \mathcal{Y}^u = \{1, \dots, C^u\}$ to the data in $D^u$.

In the specific setup of NCD, there is no overlap between the classes of $\mathcal{Y}^l$ and $\mathcal{Y}^u$, so we have $\mathcal{Y}^l \cap \mathcal{Y}^u = \emptyset$.
We are not concerned with the accuracy on the classes of $D^l$, this set is only here to provide a form of knowledge on what constitutes a relevant class.
In all the works reviewed in the paper, the number of novel classes $C^u$ is assumed to be known a priori, although we will see that some works attempt to estimate this number.
\\

\begin{figure}[htpb]
	\centering
	\includegraphics[width=0.95\linewidth]{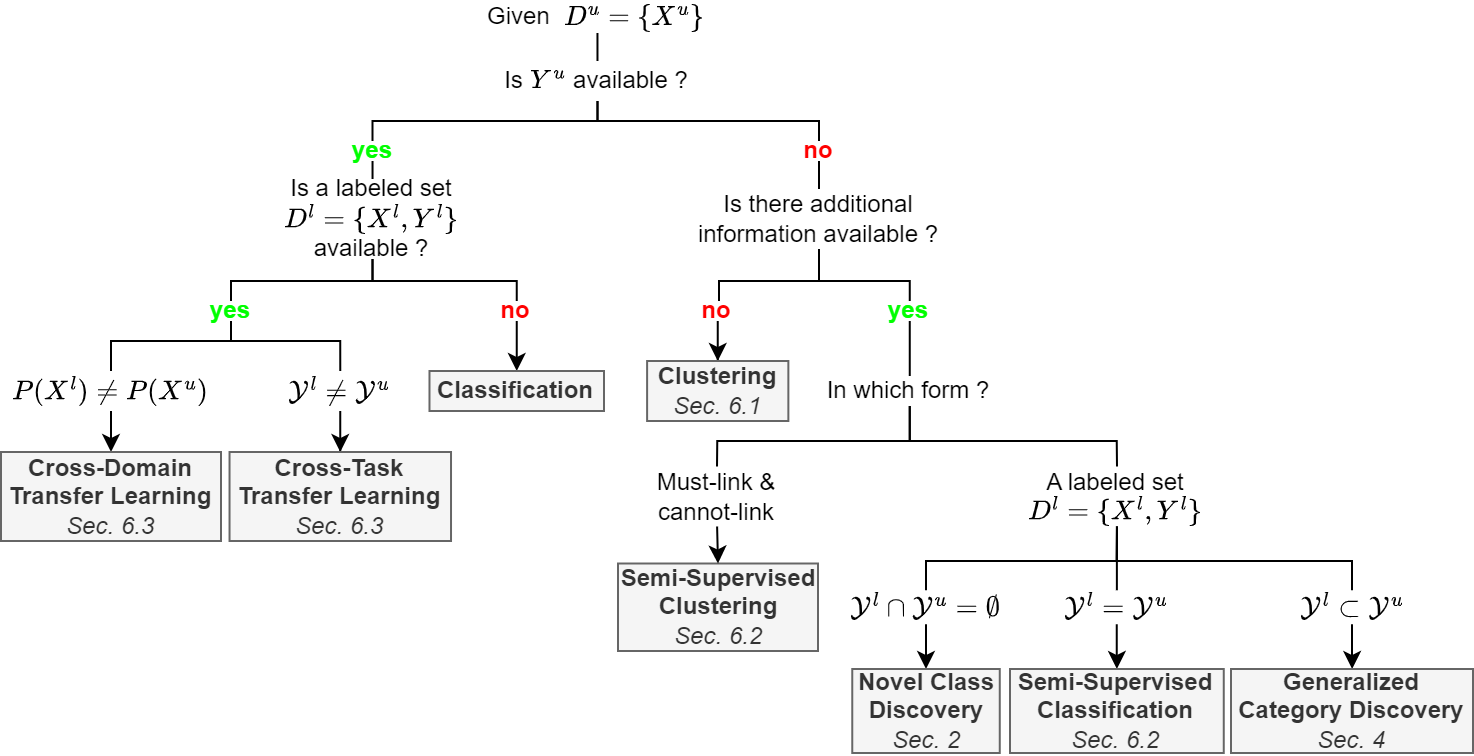}
    \caption{Overview of the domains related to Novel Class Discovery.}
    \label{fig:map_of_domains_NEW}
\end{figure}

\textbf{Positioning and key concepts of NCD:} Novel Class Discovery is a nascent and young problem with a setup that can be challenging to differentiate from other fields.
To provide an overview of the domains explored in this paper, we propose Figure~\ref{fig:map_of_domains_NEW}.
By comparing NCD with these related domains and highlighting the key differences, we aim to offer the reader a clear and comprehensive understanding of the NCD domain.
Please refer to Section~\ref{sec:ncd_related_domains} for further details and discussions.
Note that in Figure~\ref{fig:map_of_domains_NEW}, the domains are differentiated only by their setup, and while they may be similar, they dot not solve exactly the same problems.
Additionally, Open World Learning is reviewed in Section~\ref{sec:owl} but does not appear in this figure.
This is due to its broad definition and the multitude of domains it encompasses, which would cause it to appear in several branches of Figure~\ref{fig:map_of_domains_NEW}.
\\

\textbf{Evaluation protocol and metrics in NCD:} To evaluate a NCD method on a given dataset, the typical procedure \cite{han2019learning} is to hold out (or \textit{hide}) during the training phase a portion of the classes from a fully labeled dataset to act as novel classes and form the unlabeled dataset $D^u$.
For example, in most articles evaluated on MNIST, the authors consider the first 5 digits as known classes and the last 5 as novel classes whose labels are not used during training.
The performance metrics are only computed on $D^u$, as NCD is only concerned with the performance on the novel classes.

The primary metric used to evaluate the performance of models in NCD is the clustering accuracy (ACC). First introduced by \cite{yang2010image}, it requires to optimally map the predicted labels to the ground-truth labels, as the cluster numbers won't necessarily match the class numbers.
The mapping can be obtained with the the Hungarian algorithm \cite{Kuhn55hungarian} (also known as the Kuhn-Munkres algorithm). The ACC is defined as:
\begin{equation}
    ACC = \frac{1}{M} \sum_{i=1}^{M} \mathds{1}[y_i^u = \text{map}(\hat{y}_i^u)]
\end{equation}
where $\text{map}(\hat{y}_i^u)$ is the mapping of the predicted label for sample $x_i^u$ and $M$ is the number of samples in the unlabeled set $D^u$.

Another popular metric is the normalized mutual information (NMI). It measures the correspondence between the predicted and ground-truth labels and is invariant to permutations. It is defined as:
\begin{equation}
    NMI = \frac{I(\hat{y}^u, y^u)}{\sqrt{H(\hat{y}^u)H(y^u)}}
\end{equation}
where $I(\hat{y}^u, y^u)$ is the mutual information between $\hat{y}^u$ and $y^u$ and $H(y^u)$ and $H(\hat{y}^u)$ are the marginal entropies of the empirical distributions of $y^u$ and $\hat{y}^u$ respectively.

Both metrics range between 0 and 1, with values
closer to 1 indicating a better agreement to the ground truth labels.
Other metrics that can be found in NCD articles include the Balanced Accuracy (BACC) and the Adjusted Rand Index (ARI).
In the case of imbalanced class distribution, the BACC provides a more representative evaluation of the performance of a model compared to the simple accuracy.
It is calculated as the average of sensitivity and specificity.
And the ARI gives a normalized measure of agreement between the predicted clusters and the ground truth.
Unlike the other metrics, it ranges from -1 to 1, with higher values also indicating better agreement between the two clusterings.
A score of 0 indicates random clustering, while negative scores indicate a performance worse than random.

\section{Taxonomy of Novel Class Discovery methods}
\label{sec:taxonomy}

In this section, NCD works are organized by the way in which they transfer knowledge from the labeled set $D^l$ to the unlabeled set $D^u$.
Also identified by \cite{liu2022residual}, and \cite{2204.10595}, NCD methods adopt either a \textit{one-} or \textit{two-stage} approach.
An overview of the methods that are studied in this section is provided in Table~\ref{tab:ncd_contributions}, along with a brief description of their contributions.

The first NCD works published were generally two-stage approaches, so they are described here first.
They tackle the NCD problem in a way similar to cross-task Transfer Learning (TL) methods.
They first focus on $D^l$ only (like a source dataset in TL) before exploring $D^u$ (similarly to a target dataset without labels in TL).
Within this category, two families of methods can be distinguished:
one uses $D^l$ to learn a similarity function, while the other incorporates the features relevant to the classes of $D^l$ into a latent representation.

More recent methods adopt one-stage approaches and process $D^l$ and $D^u$ simultaneously through a shared objective function.
All the one-stage methods reviewed here work in a similar manner, where a latent space shared by $D^l$ and $D^u$ is trained by two classification networks with different objectives.
These objectives usually include clustering the unlabeled data and maintaining good classification accuracy on the labeled data.

\begin{table}[htpb]
    \centering
    \fontsize{7}{8}\selectfont
    \begin{tblr}{hline{1-Z} = {solid}, vline{1-Z}={solid},
                colspec = { m{2.5mm} >{\RaggedRight}p{23mm} l m{70mm} }}
        \SetCell[c=2]{c}{\textbf{Knowledge}\\\textbf{transfer method}} & & \SetCell[c=1]{c}{\textbf{Article}} & \SetCell[c=1]{c}{\textbf{Main contributions}} \\
        \SetCell[r=4]{}{\rotatebox[origin=c]{90}{Two-stage methods}}
            & \SetCell[r=2]{}{Similarity function learned on $D^l$}
                & CCN \cite{hsu2018learning} & The first article to define and solve the NCD problem. \\
                \cline{3-4}
            &   & MCL \cite{hsu2019multiclass} & Improvement of \cite{hsu2018learning} and introduction of the modified binary cross-entropy with inner product. \\
            \cline{2-4}
            & \SetCell[r=2]{}{Latent space learned on $D^l$}
                & DTC \cite{han2019learning} & Adaptation of a deep clustering method \cite{xie2016unsupervised} for NCD. \\
                \cline{3-4}
            &   & MM/MP \cite{chi2022meta} & Formalization of the assumptions behind NCD. Solving NCD with a limited quantity of unlabeled data. \\ 
        \SetCell[r=9]{}{\rotatebox[origin=c]{90}{One-stage methods}}
            & \SetCell[r=9]{}{Joint objective on $D^l$ and $D^u$}
                & AutoNovel \cite{autonovel2021, han2020automatically} & Using SSL to pre-train using all the data. The RankStats method for pseudo labeling. Joint objective of classification on $D^l$ and clustering on $D^u$. \\
                \cline{3-4}
            &   & CD-KNet-Exp \cite{wang2020openworld} & Using the Hilbert Schmidt Independence Criterion to bridge supervised and unsupervised information. \\
                \cline{3-4}
            &   & \textit{Unnamed} \cite{qing2021end} & Insertion of the pre-training objective in the joint loss. \\
                \cline{3-4}
            &   & OpenMix \cite{zhong2020openmix} & Creating synthetic samples with mixed known and unknown classes to produce robust pseudo labels. \\
                \cline{3-4}
            &   & NCL \cite{zhong2021neighborhood} & Adapting contrastive learning to the NCD setting, along with NCD-specific hard-negative generation. \\
                \cline{3-4}
            &   & WTA \cite{2104.12673} & A solution for NCD in multi-modal video data, using WTA hashing \cite{yagnik2011power} for pseudo labeling. \\
                \cline{3-4}
            &   & DualRS \cite{zhao2021novel} & Automatic extraction of both global and local features of images to define robust pseudo labels.\\
                \cline{3-4}
            &   & Spacing loss \cite{2204.10595} & Learning an easily separable representation with spaced-out spherical clusters. \\
                \cline{3-4}
            &   & TabularNCD \cite{tabularncd} & Solving the NCD problem for tabular datasets.\\
    \end{tblr}
    \caption{Main contributions of the works in NCD, organized by the method of knowledge transfer from $D^l$ to $D^u$.}
    \label{tab:ncd_contributions}
\end{table}

\subsection{Two-stage methods}
\label{sec:two_stage_methods}

\subsubsection{Learned-similarity--based}
\label{sec:learned_similarity_based}

\begin{figure}[H]
	\centering
	\includegraphics[width=0.95\linewidth]{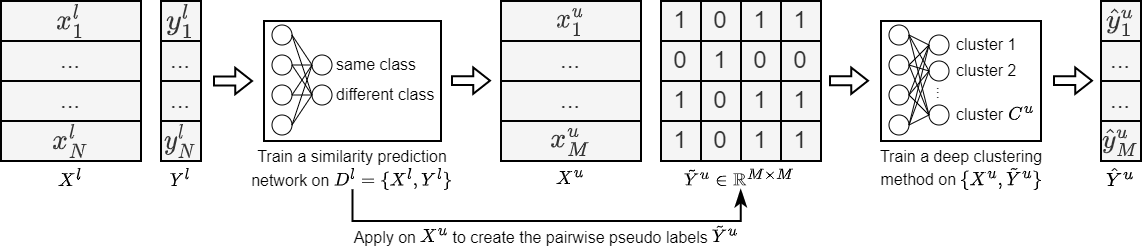}
    \caption{General workflow of learned-similarity--based methods.}
    \label{fig:similarity_based}
\end{figure}

The general workflow of learned-similarity–based methods is illustrated in Figure~\ref{fig:similarity_based}.
Learned-similarity--based methods start by learning on $D^l$ a function that is also applicable on $D^u$ and determines if pairs of instances belong to the same class or not.
As the numbers $C^l$ and $C^u$ of classes can be different, a \textit{binary} classification network is generally trained by deriving supervised pairwise labels from the existing class labels $Y^l$.
The learned binary classifier is then applied on each unique pair of instances in the unlabeled set $D^u=\{X^u\}$ to form a pairwise pseudo label matrix $\Tilde{Y}^u$.
This matrix is used as a target to train a classifier on $D^u$ and make the final class prediction. 

In this section, we review two of the main learned-similarity–-based methods of the literature.
CCN \cite{hsu2018learning} is the first to tackle the very specific problem of NCD, and MCL \cite{hsu2019multiclass} makes improvements to CCN and defines a loss function used in many subsequent NCD works.
\\

\textbf{$\bullet$ Constrained Clustering Network (CCN)} \cite{hsu2018learning} tackles the cross-domain Transfer Learning (TL) problem which is outside of the scope of this review, as well as a cross-task TL problem that corresponds to NCD.
In the latter, the method seeks to cluster $D^u$ by using the knowledge of a network trained on $D^l$.
In the first stage, a similarity prediction network is trained on $D^l$ to distinguish if pairs of instances belong to the same class or not.
This network is then applied on $D^u$ to create a matrix of pairwise pseudo labels $\Tilde{Y}^u$ (similarly to must-link and cannot-link constraints).
In the second stage, a new classification network is defined with $C^u$ output neurons with the objective of partitioning $D^u$.
It is trained on $D^u$ by comparing the previously defined pseudo labels to the KL-divergence between pairs of its cluster assignments.
In other words, if for two samples $x_i$ and $x_j$ the value in the pseudo labels matrix is 1 (i.e. $\Tilde{Y}^u_{i,j}=1$), the two cluster assignments of the classification network must match according to the KL-divergence.
The idea behind this approach is that if a pair of instances is similar, then their output distribution should be similar (and vice-versa), resulting in clusters of similar instances according to the similarity network.
%
\\
\\
\indent\textbf{$\bullet$ Meta Classification Likelihood (MCL)} \cite{hsu2019multiclass} is a continuation of CCN \cite{hsu2018learning} by the same authors.
They also consider multiple scenarios, one of them being ``unsupervised cross-task transfer learning'', which corresponds to the NCD setting.
Similarly to CCN \cite{hsu2018learning}, pairwise pseudo labels are constructed on $D^u$ by a similarity prediction network trained on $D^l$.
A classification network with $C^u$ output neurons is also defined to partition $D^u$.
But this time, the KL-divergence is not used to determine if two instances were assigned to the same class.
Instead, they use the inner product of the prediction $p_{i,j} = \hat{y}_{i}^T \cdot \hat{y}_{j}$.
This $p_{i,j}$ will be close to 1 when the predicted distributions $\hat{y}_{i}$ and $\hat{y}_{j}$ are sharply peaked at the same output node and close to 0 otherwise.
This is a simple yet effective idea that can be directly compared to the pairwise pseudo labels $\Tilde{y}_{i,j} \in \{0,1\}$ and enables the use of the usual binary cross-entropy (BCE) as a loss function:
\begin{equation}
    L_{BCE} = -\sum_{i,j}\Tilde{y}_{i,j} \text{log}(\hat{y}_{i}^T \cdot \hat{y}_{j}) + (1 - \Tilde{y}_{i,j}) \text{log}(1 - \hat{y}_{i}^T \cdot \hat{y}_{j})
    \label{eq:modified_bce}
\end{equation}
This is an important formalization of the classification problem with pairwise labels that has been used in many subsequent NCD papers.

\subsubsection{Latent-space--based}

\begin{figure}[H]
	\centering
	\includegraphics[width=0.95\linewidth]{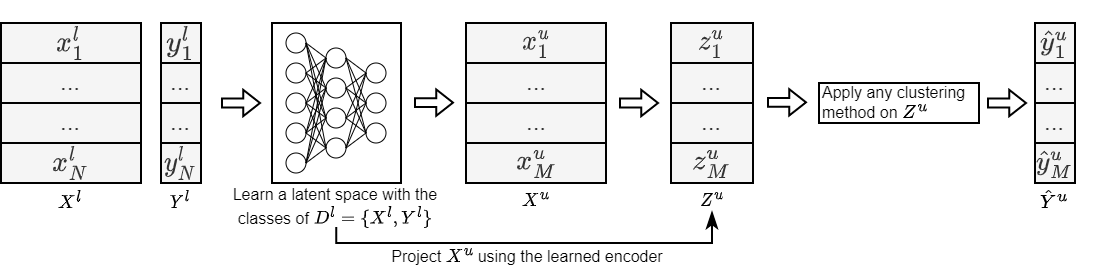}
    \caption{General workflow of latent-space--based methods}
    \label{fig:latent_space_based}
\end{figure}

The general workflow of latent-space--based methods is illustrated in Figure~\ref{fig:latent_space_based}.
These methods start by training with $D^l=\{X^l,Y^l\}$ a latent representation that incorporates the important characteristics of the known classes $\mathcal{Y}^l$.
This is usually done by defining a deep classifier with several hidden layers.
After training with cross-entropy, the output and softmax layers are discarded, and the last hidden layer is now regarded as the output of an encoder.
These methods make the assumption that the high-level features of the known classes are shared by the unknown classes.
As the latent space highlights these features, $X^u$ is then projected inside, and any off-the-shelf clustering method can be applied to discover the unknown classes.

Two relevant latent-space--based methods are summarized below.
DTC \cite{han2019learning} extends to the NCD setting a deep clustering method, which is very suitable for the NCD problem.
MM \cite{chi2022meta} formalizes the assumptions behind NCD and proposes to train a set of expert classifiers to cluster the unlabeled data.
\\

\textbf{$\bullet$ Deep Transfer Clustering (DTC)} \cite{han2019learning} is based on an unsupervised deep clustering method, DEC \cite{xie2016unsupervised}, which clusters the data while learning a good representation at the same time.
Unlike many deep clustering methods, DEC does not rely on pairwise pseudo labels. Instead, it maintains a list of class prototypes that represent the cluster centers and assigns instances to the closest prototype.
To adapt DEC to the NCD setting, DTC initializes a representation by training a classifier with cross-entropy on $D^l$ using the ground truth labels. The embedding of $D^u$ is then obtained by projecting through the classifier whose last layer was removed.
An intuitive conclusion for DEC is that if the classes $Y^l$ and $Y^u$ share similar semantic features, DEC should perform better on the embedding of $D^u$ produced this way.

After projection of $D^u$, DTC applies DEC with some improvements.
Namely, the clusters are slowly annealed to prevent collapsing the representation to the closest cluster centers, and they find that further reducing the dimension of the learned representation with Principal Component Analysis (PCA) leads to an improved performance.
\\
\\
\indent\textbf{$\bullet$ Meta Discovery with MAML (MM)} \cite{chi2022meta} proposes a new method along with theoretical contributions to the field of NCD, by defining a set of conditions that must be met so that NCD is theoretically solvable.
In simple terms, they state that:
(1) known and novel classes must be disjoint
(2) it must be meaningful to separate observations from $X^l$ and $X^u$
(3) good high-level features must exist for $X^l$ or $X^u$
and based on these features, it must be easy to separate $X^l$ or $X^u$
(4) these high-level features are shared by $X^l$ and $X^u$.
These four conditions are worthy of consideration when the NCD problem is addressed for a new dataset.
The reader may find more details in the original article.

Based on the assumption that $X^l$ and $X^u$ share high level features where the partitioning is easy, the authors suggest that it is possible to cluster $D^u$ based on the features learned on $D^l$.
Therefore, they propose a two-stage approach that starts by training a number of ``expert'' classifiers on $D^l$ with a shared feature extractor.
These classifiers are constrained to be orthogonal to each other to ensure that they each learn to recognize unique features of the labeled data.
The resulting latent space should reveal these high-level features, shared by the labeled and unlabeled data, and should be sufficient to cluster $D^u$.
The expert classifiers are then fine-tuned on the unlabeled data $D^u$ with the BCE of Equation~\eqref{eq:modified_bce} by defining pseudo labels based on the similarity of instances in the latent representation learned on $D^l$.
The output of the classifiers after fine-tuning is used as the final prediction for the unlabeled data.

This paper also makes experiments given a limited quantity of unlabeled data, and shows that its method is more robust than the competitors in this case.

\subsection{One-stage methods}
\label{sec:one_stage_ncd}

\subsubsection{Introduction}
\begin{figure}[H]
	\centering
	\includegraphics[height=0.21\linewidth]{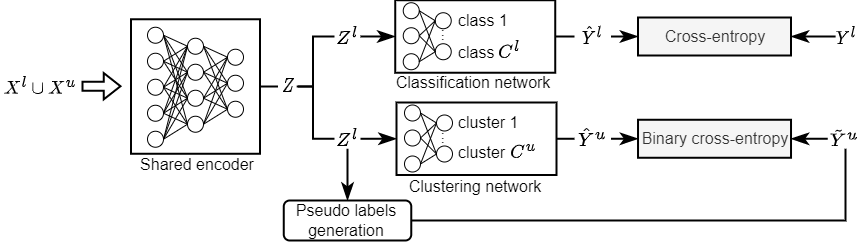}
    \caption{General workflow of one-stage methods. The regularization loss is omitted for the sake of clarity.}
    \label{fig:one_stage_methods}
\end{figure}

The general workflow of one-stage methods is illustrated in Figure~\ref{fig:one_stage_methods}.
In opposition to two-stage methods, one-stage methods exploit both sets $D^l$ and $D^u$ simultaneously.
Some of these methods still have multiple steps (such as pre-training on $D^l$), but they are characterized by their joint use of $D^l$ and $D^u$ during the clustering phase.
Among two-stage approaches, both similarity (see Section \ref{sec:two_stage_methods}.A) and latent-space based (see Section \ref{sec:two_stage_methods}.B) are negatively impacted when the relevant high-level features are not completely shared by the known and unknown classes, as shown in \cite{li2022a}. But by handling data from both sets of classes, one-stage methods will inherently obtain a better latent representation less biased towards the known classes.

Most one-stage methods jointly train two classification networks (see Figure~\ref{fig:one_stage_methods}). One predicts the labels of $D^l$ and introduces the relevant features of the known classes, and the other partitions $D^u$ using pseudo labels usually defined with similarity measures.
By training both networks on the same latent space, they share knowledge with other.
In this survey, the classification network trained on $D^u$ will be referred to as a ``clustering'' network, since it is trained with unlabeled data.

One-stage methods define a multi-objective loss function which typically has 3 components:
cross-entropy ($\mathcal{L}_{CE}$), binary cross-entropy ($\mathcal{L}_{BCE}$) and regularization ($\mathcal{L}_{MSE}$).
The cross-entropy loss is simply used to train the classification network with the ground-truth labels.
The binary cross-entropy loss compares the prediction of the clustering network to pseudo labels (see Equation~\eqref{eq:modified_bce}).
And the regularisation loss ensures that the model generalizes to a good solution.
This is usually done by encouraging both networks to predict the same class for an instance and its randomly augmented counterpart (see column ``Data Augmentation'' in Table~\ref{tab:methods_characteristics}).

While Section~\ref{sec:two_stage_methods} was, to the best of our knowledge, an exhaustive list of the two-stage methods, there is a larger (and fast growing) number of papers that follow a one-stage approach.
For this reason, only four methods representative of the literature are first detailed, and a few other methods are described more concisely in the last section.

\subsubsection{AutoNovel}
AutoNovel \cite{autonovel2021, han2020automatically} is the first one-stage method proposed to solve the NCD problem.
It introduced the architecture illustrated in Figure~\ref{fig:one_stage_methods} and inspired many subsequent works \cite{zhong2021neighborhood, 2204.10595, zhong2020openmix, 2104.12673, zhao2021novel}.
AutoNovel starts by carefully initializing its encoder using the RotNet \cite{gidaris2018unsupervised} Self-Supervised Learning (SSL) method to train on both labeled and unlabeled data.
As SSL does not leverage the labels of known classes, the learned features will not be biased towards the known classes.
At this point, the authors consider that the features learned by the encoder will be representative of all data and will be useful for any given task, so they \textit{freeze} all but the last layer of the encoder.
Finally, the labeled data is used to train for a few epochs the classifier and fine-tune the last layer of the encoder.
This concludes the initialization of the representation (the shared encoder in Figure~\ref{fig:one_stage_methods}), which is crucial as the next step involves determining pseudo labels in the latent space based on pairwise similarity measures.

To realize the joint learning on $D^l$ and $D^u$, the two classification networks that can be seen in Figure~\ref{fig:one_stage_methods} are added on top of the encoder.
The three components of the model (shared encoder, classification network and clustering network) are then trained using a loss composed of the three components described in the introduction of this section:
\begin{equation}
    \mathcal{L}_{AutoNovel} = \mathcal{L}_{CE} + \mathcal{L}_{BCE} + \mathcal{L}_{MSE}
    \label{eq:autonovel}
\end{equation}
As AutoNovel uses the BCE of Equation~\eqref{eq:modified_bce}, the inner products of the clustering network predictions are compared to the pairwise pseudo labels defined by their original RankStats (for \textit{ranking statistics}) method (see Section~\ref{sec:pseudo_label_def}).

\subsubsection{Class Discovery Kernel Network with Expansion (CD-KNet-Exp)}
CD-KNet-Exp \cite{wang2020openworld} is a multi-stage method that constructs a latent representation using $D^l$ and $D^u$ that is suitable, after training, to the discovery of the novel classes by a $k$-means.
It starts by pre-training a representation with a ``deep'' classifier on $D^l$ only.
Since this embedding could be highly biased towards the known classes, and may not generalize well to $D^u$, the representation is then fine-tuned with both $D^l$ and $D^u$.
In this second stage, they optimize the following objective:
\begin{equation}
    \underset{U, \theta}{\text{max}} \text{ } \mathbb{H}(f_\theta(X), U) + \lambda \mathbb{H}(f_\theta(X^l), Y^l)
\end{equation}
$f$ is the feature extractor (or encoder) of parameter $\theta$. $\mathbb{H}(P, Q)$ is the Hilbert Schmidt Independence Criterion (HSIC). It measures the dependence between distributions $P$ and $Q$. And $U$ is the spectral embedding of $X$. Intuitively, the first term encourages the separation of all classes (old and new) by performing something similar to spectral clustering. And the second term introduces the supervised information from the known classes by maximizing the dependence between the embedding of $X^l$ and its labels $Y^l$.

This second step produces a latent space that should have incorporated the information from both known and unknown classes and be easily separable.
For this reason, the embedding of the data is finally $f_\theta(X^u)$ partitioned with $k$-means clustering.

\subsubsection{OpenMix}
The principle of OpenMix \cite{zhong2020openmix} is to exploit the labeled data to generate more robust pseudo labels for the unlabeled data.
It relies on MixUp \cite{zhang2018mixup}, which is widely used in supervised and semi-supervised learning.
As MixUp requires labeled samples for every class of interest, applying it directly on the unlabeled data would still produce unreliable pseudo labels.
Instead, OpenMix generates new training samples by mixing both labeled and unlabeled samples.

\begin{figure}[H]
	\centering
	\includegraphics[height=0.15\linewidth]{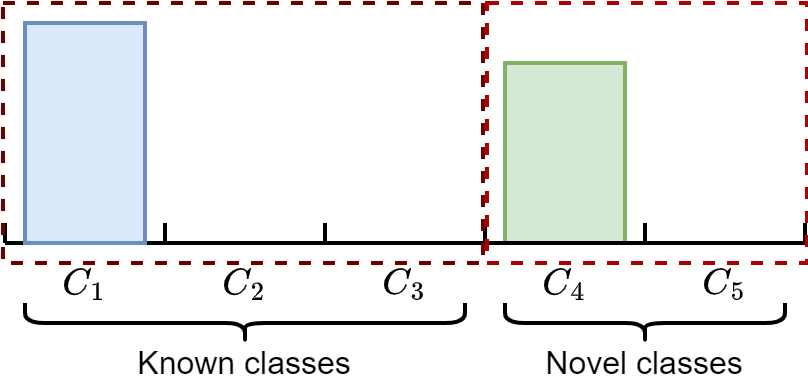}
    \caption{Example of synthetic label generated by Openmix \cite{zhong2020openmix}. Here, it is a mix of a labeled sample of class $C_1$ and an unlabeled sample with pseudo label $C_4$.}
    \label{fig:open_mix_labels}
\end{figure}

First, a latent representation is initialized using the known classes only.
Then, a clustering network is defined to discover the new classes using a joint loss on $D^l$ and $D^u$.
The model is trained with synthetic data that are a mix of a sample from a \textit{known class} and a sample from an \textit{unknown class}.
The synthetic data points are generated with MixUp, while the labels are a combination of the ground-truth labels of the labeled samples and the pseudo labels determined using cosine similarity for the unlabeled samples (see Figure~\ref{fig:open_mix_labels}).
The authors argue that the overall uncertainty of the resulting pseudo labels will be reduced, as the labeled counterpart does not belong to any new class and its label distribution is exactly true.

These synthetic labels are compared to the prediction of the model: (i) the classification network predicts the known part and (ii) the clustering network the unknown part (see Figure~\ref{fig:one_stage_methods}) of the full label space.

The authors observe that the clustering network has good accuracy on the samples that it predicted with high-confidence.
Based on this observation, they regard these samples as \textit{reliable anchors} that are further integrated with unlabeled samples to generate even more combinations with MixUp.

\subsubsection{Neighborhood Contrastive Learning (NCL)}
\label{sec:ncl}
NCL \cite{zhong2021neighborhood} is inspired by AutoNovel \cite{autonovel2021} as it uses the same architecture (see Figure~\ref{fig:one_stage_methods}) and pre-trains its representation in the same way.
Its main contribution is the addition of 2 contrastive learning terms to the loss of AutoNovel (see Equation~\eqref{eq:autonovel}) to improve the learning of discriminative representations.
The first one is the supervised contrastive learning term from \cite{khosla2021supervised} applied to the labeled data using the ground-truth labels.
The second term is applied on the unlabeled data and adapts the original unsupervised contrastive learning loss to the NCD problem to exploit both labeled and unlabeled data.

For this second term, the authors maintain a queue $M^u$ of samples from past training steps, and consider for any instance in a batch that the $k$ most similar instances from the queue are most likely from the same class. The contrastive loss, for these \textit{positive} pairs is defined for the embedding $z_i^u$ of an instance $x_i^u$ as:
\begin{equation}
    l(z^u_i, \rho_k) = - \frac{1}{k} \sum_{\Bar{z}_j^u \in \rho_k} \text{log} \frac{e^{\delta(z^u_i, \Bar{z}^u_j)/\tau}}{e^{\delta(z^u_i, \hat{z}^u_i)/\tau} + \sum_{m=1}^{|M^u|} e^{\delta(z^u_i,\Bar{z}^u_m)/\tau}}
    \label{eq:ncl2}
\end{equation}
with $\rho_k$ the $k$ instances most similar to $z^u_i$ in the unlabeled queue $M^u$, $\delta$ the similarity function and $\tau$ a temperature parameter.

Additionally, synthetic positive pairs $(z^u, \hat{z}^u)$ are generated by randomly augmenting each instance. The contrastive loss for positive pairs is written as:
\begin{equation}
    l(z^u, \hat{z}^u) = - \text{log} \frac{e^{\delta(z^u, \hat{z}^u)/\tau}}{e^{\delta(z^u, \hat{z}^u)/\tau} + \sum_{m=1}^{|M^u|} e^{\delta(z^u,\Bar{z}^u_m)/\tau}}
    \label{eq:ncl1}
\end{equation}

Finally, ``hard negatives'' are introduced in the queue $M^u$ to further improve the learning process.
Hard negatives refer to similar samples that belong to a different class and are an important concept in contrastive learning.
Selecting hard negatives in $D^u$ can be difficult since there are no class labels available.
Therefore, the authors take advantage of the fact that the classes of $D^l$ and $D^u$ are necessarily disjoint and create new hard negative samples by interpolating easy negatives from the unlabeled set (i.e. instances that are most likely true negatives) with hard negatives from the labeled set.

To summarize, the overall loss that is optimized by the model is:
\begin{equation}
    \mathcal{L}_{NCL} = \mathcal{L}_{AutoNovel} + l_{scl} + \alpha l(z^u_i, \rho_k) + (1 - \alpha) l(z^u, \hat{z}^u)
\end{equation}
where $l_{scl}$ is the supervised contrastive loss term for the labeled samples of $D^l$ and $\alpha$ is a trade-off parameter.


\subsubsection{Other methods}
We briefly describe a few other one-stage NCD works here.
In \cite{qing2021end}, the SSL objective of RotNet \cite{gidaris2018unsupervised} and joint objective of Equation~\eqref{eq:autonovel} are merged in a single loss function.
The shared encoder is therefore influenced by the classification network, the clustering network and a linear layer that predicts the random rotations of images.
The authors argue that the self-supervised signals will provide a strong regularization that will alleviate the performance degradation caused by the noisy pseudo labels.

The method proposed in \cite{2104.12673} is able to process multi-modal data, composed of both video and audio.
Two feature encoders are trained with Noise Contrastive Estimation (NCE) \cite{gutmann2010noise}, and the latent representations are concatenated before being fed to either a classification or clustering network.
The Winner-Take-All hash \cite{yagnik2011power} is used to measure the similarity between each pair of unlabeled samples during the definition of pseudo labels required to train the clustering network.
The authors argue that WTA is more robust to noise and effectively captures the structural relationships among the objects (see \cite{2104.12673} for more details).

The Dual Ranking Statistics (DualRS) \cite{zhao2021novel} method trains two framework branches on a shared latent representation.
Both branches have a classifier trained to predict the known classes and a clustering network trained with pseudo labels and Equation~\eqref{eq:modified_bce}.
One branch is tasked to extract global features, as pseudo labels are defined by measuring the similarity between \textit{whole} images.
The other branch focuses on individual local details, and pairwise similarities are computed using only part of each image.
The authors argue that these branches are complementary to each other, as they focus on different granularity of the data. The global branch may easily find similarities and introduce more false positives and have high recall (but low precision), while the local-part branch will be more ``strict'' and have high precision (but low recall).
To make the two branches communicate, agreement between the similarity score distributions of unlabeled data is encouraged.

Similarly to \cite{wang2020openworld}, the Spacing Loss \cite{2204.10595} method shapes a latent space where the novel classes are easily separable.
During training, the representation is slowly guided to have spaced-out clusters that are equidistant to each other.
Each epoch alternates between learning with pseudo labels derived from the closest cluster centers and modifying the cluster centers themselves.
During inference, a $k$-means is run in the learned latent representation to discover the novel categories.

Finally, to the best of our knowledge, a single method has attempted to solve NCD in the context of tabular data \cite{tabularncd}.
It pre-trains a simple encoder of dense layers with the VIME \cite{vime} self-supervised learning method and adopts the two heads architecture of Figure~\ref{fig:one_stage_methods}.
Similar to other one-stage methods, known classes are classified jointly with clustering on the unlabeled data, and pseudo labels are defined based on pairwise cosine similarity.

\subsection{Estimating the number of unknown classes}

The assumption that the number $C^u$ of unknown classes in the unlabeled set $D^u$ is known can be unrealistic in some scenarios.
For this reason, a few methods were proposed to automatically estimate this number $C^u$.

A method used in \cite{hsu2018learning, hsu2019multiclass, zhao2021novel, yu2022self}, consists in setting the number of output neurons of the clustering network to a large number (e.g. 100).
In doing so, we rely on the clustering network to use only the necessary number of clusters and to leave the other output neurons unused.
Clusters are counted if they contain more instances than a certain threshold.
This approach is surprisingly simple, but displays stable results in the different articles that experimented it.

In \cite{vaze2022generalized, fei2022xcon}, a $k$-means is performed on the entire dataset $D^l \cup D^u$.
The number of unknown classes $C^u$ is estimated to be the $k$ that maximized the Hungarian clustering accuracy (see Section~\ref{sec:preliminaries}): a $k$ too high will result in clusters assigned to the null set and a number too low will have clusters composed of multiple classes, both cases will be considered as being assigned incorrectly.

\begin{figure}[htbp]
	\centering
	\includegraphics[width=0.6\linewidth]{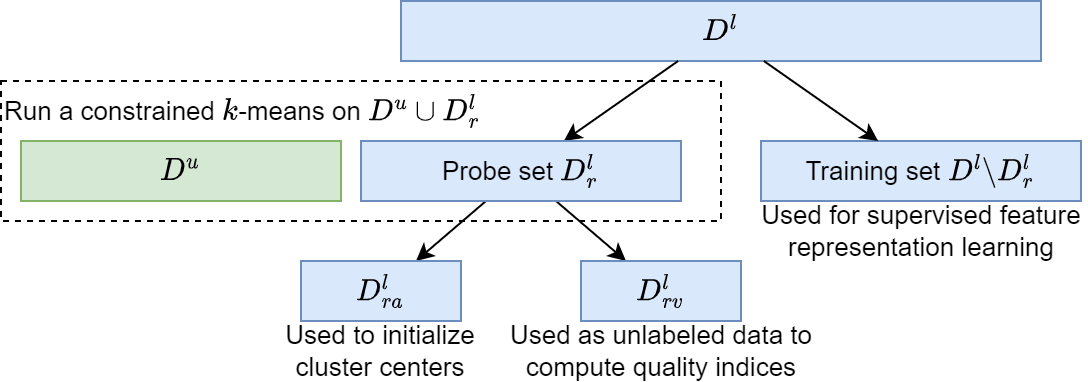}
    \caption{Number of unknown classes estimation process from DTC \cite{han2019learning}.}
    \label{fig:k_clusters_estimation}
\end{figure}

Finally, another popular idea is to make use of the known classes \cite{han2019learning, han2020automatically, autonovel2021, zheng2022openset}.
This process is illustrated in Figure~\ref{fig:k_clusters_estimation}.
The known classes of $D^l$ are first split into a \textit{probe} subset $D^l_r$ and a training subset $D^l \backslash D^l_r$ containing the remaining classes.
The set $D^l \backslash D^l_r$ is used for supervised feature representation learning, while the probe set $D^l_r$ is combined with the unlabeled set $D^u$.
Now, a constrained $k$-means is run on $D^l_r \cup D^u$.
Part of the classes of $D^l_r$ are used for the clusters initialization, while the rest are used to compute 2 cluster quality indices (average clustering accuracy and cluster validity index, see \cite{han2019learning}).
Note that this can be difficult to use when the number of known classes is small, since it involves many class splits.

\subsection{Methods summary}

Table~\ref{tab:methods_characteristics} summarizes the important characteristics of the methods that were described in this section.
These characteristics include the type of data processed, the method of defining pairwise pseudo labels and, if applicable, the method of estimating the number of unknown classes $C^u$.
From column ``Unknown $C^u$'', it is evident that all the works reviewed here assume knowledge of the number of unknown classes.
Moreover, this table highlights the popularity of pairwise pseudo labeling as a means of training classification networks on unlabeled data, with only DTC \cite{han2019learning} and CD-KNet-Exp \cite{wang2020openworld} relying on different processes.

\begin{table}[htpb]
    \fontsize{7}{8}\selectfont
    \centering
    \begin{tblr}{hline{1} = {2-Z}{solid}, vline{1} = {2-Z}{solid},
                hline{2-Z} = {solid}, vline{2-Z}={solid},
                colspec = { m{2.5mm} l >{\centering\arraybackslash}m{11mm} >{\centering\arraybackslash}m{17mm} >{\centering\arraybackslash}m{19mm} >{\centering\arraybackslash}m{12.5mm} >{\centering\arraybackslash}m{17mm} >{\centering\arraybackslash}m{14mm} }}
            & \SetCell[c=1]{c}{\textbf{Method}} & \textbf{Data Type} & \textbf{Backbone architecture} & \textbf{Pairwise pseudo labels} & \textbf{Pre-training} & \textbf{Data Augmentation} & \textbf{Unknown $C_u$} \\
            \SetCell[r=4]{}{\rotatebox[origin=c]{90}{Two-stage methods}}
                & CCN \cite{hsu2018learning}                           & Image & ResNet18 & From learned classifier & \ding{55} & \ding{55} &\ding{55} + Estimated ($k=100$) \\
                & MCL \cite{hsu2019multiclass}                         & Image & LeNet, VGG8 and ResNet & From learned classifier & \ding{55} & Crop and flip &\ding{55} + Estimated ($k=100$) \\
                & DTC \cite{han2019learning}                           & Image & ResNet18 and VGG & \ding{55} (class prototypes) & CE on $D^l$ & Crop and flip & \ding{55} + Estimated (probe classes) \\
                & MM/MP \cite{chi2022meta}                             & Image & ResNet18 and VGG16 & RankStats \cite{autonovel2021} & CE on $D^l$ & \ding{55} & \ding{55} \\
            \SetCell[r=9]{}{\rotatebox[origin=c]{90}{One-stage methods}}
                & AutoNovel \cite{autonovel2021, han2020automatically} & Image & VGG and ResNet18 & RankStats \cite{autonovel2021} & RotNet \cite{gidaris2018unsupervised} on $D^l \cup D^u$ & Crop and flip & \ding{55} + Estimated (probe classes) \\
                & CD-KNet-Exp \cite{wang2020openworld}                 & Image & Custom CNN & \ding{55} & CE on $D^l$ & \ding{55} & \ding{55} \\
                & \textit{Unnamed} \cite{qing2021end}                  & Image & ResNet18 & Threshold on SNE & \ding{55} & Yes, unspecified & \ding{55} \\
                & OpenMix \cite{zhong2020openmix}                      & Image & VGG and ResNet18 & Threshold cosine similarity & CE on $D^l$ & Crop and flip & \ding{55} \\
                & NCL \cite{zhong2021neighborhood}                     & Image & ResNet18 & Threshold cosine similarity & RotNet \cite{gidaris2018unsupervised} on $D^l \cup D^u$ & Crop and flip & \ding{55} \\
                & WTA \cite{2104.12673}                                & {Image \& \\ Video} & R3D-18 and ResNet18 & WTA hash \cite{yagnik2011power} & \ding{55} & Crop, resize, flip, color distortion and blur & \ding{55} \\
                & DualRS \cite{zhao2021novel}                          & Image & RestNet18 & Dual ranking statistics & RotNet \cite{gidaris2018unsupervised} on $D^l \cup D^u$ & Crop and flip & \ding{55} + method from DTC \\
                & Spacing Loss \cite{2204.10595}                       & Image & ResNet18 & Threshold cosine sim. + class prototypes & CE on $D^l$ & Crop and flip & \ding{55} \\
                & TabularNCD \cite{tabularncd}                         & Tabular & Custom DNN & Number of most similar & VIME \cite{vime} on $D^l \cup D^u$ & SMOTE \cite{SMOTE} & \ding{55} \\
        \end{tblr}
        \caption{Overview of the characteristics of NCD methods.}
        \label{tab:methods_characteristics}
\end{table}

\section{New domains derived from Novel Class Discovery}
\label{sec:gcd}

As the number of NCD works increases, new domains closely related to it are emerging.
Researchers are designing scenarios where they relax some of the hypotheses or define new tasks inspired by NCD.
This section will provide a brief overview of some of the most important of these domains.
Given their similarity in settings, Table~\ref{tab:domains_related_to_ncd} highlights some of the key differences among them.

\begin{table}[H]
    \centering
    \begin{tabular}{|m{31mm}|c|c|c|} 
        \cline{2-4}
        \multicolumn{1}{l|}{} & NCD & GCD & NCDwF \\ 
        \hline
        test data $\in \mathcal{Y}^l \cup \mathcal{Y}^u$ & \ding{55} & \ding{51} & \ding{51} \\ 
        \hline
        $D^l$ and $D^u$ are available simultaneously & \ding{51} & \ding{51} & \ding{55} \\
        \hline
    \end{tabular}
    \caption{Distinctions between the related domains.}
    \label{tab:domains_related_to_ncd}
\end{table}

\textbf{Generalized Category Discovery (GCD)} \cite{vaze2022generalized} is a setting that is gaining traction from the community, with some very recent articles published \cite{vaze2022generalized, zheng2022openset, Yang_2022_CVPR, fei2022xcon}.
GCD was designed to be a less constrained and more realistic setting of Novel Class Discovery, as it does not assume that samples during inference will only belong to the unknown classes.
As the test data can belong to either known or unknown classes, the task at inference becomes to (i) accurately classify samples from known classes and (ii) find the clusters of samples from unknown classes.
Compared to NCD, this poses a greater challenge for designing an efficient model.
Methods in this domain are thus evaluated for both their classification and clustering performance.
Note that this setting is close to Open World Learning, but still different as the training data is still composed of two separate sets ($D^l$ and $D^u$).

This problem was first solved in 2021 by \cite{2108.08536}, but it was not immediately recognized as a setting distinct from NCD.
Later, as multiple articles were published simultaneously, different names were used and problem was presented in varying ways.
Some of these names include \textit{Generalized Novel Class Discovery} \cite{Yang_2022_CVPR}, \textit{Open Set Domain Adaptation} \cite{zhuang2022discovering} and \textit{Open-World Semi-Supervised Learning} \cite{2207.02261}, however, they all ultimately aimed to solve the same task.

In the first article that formalizes the GCD problem \cite{vaze2022generalized}, the authors find that existing NCD methods are prone to overfitting on the known classes.
Instead of using a parametric classifier, which was seemingly the cause of the overfitting, they use contrastive learning and a semi-supervised $k$-means to recognize images.

Another method of interest is XCon \cite{fei2022xcon}. In this case, the authors focus on fine-grained Generalized Category Discovery, where different classes have very close high-level features (e.g. two different species of birds where only the beak is different). They propose to partition the data into $k$ sub-datasets that share irrelevant cues (e.g. background and object pose) to force the method to focus on important discriminative information.


\textbf{Note:} GCD and its links to Open-World Learning are discussed in Section~\ref{sec:owl}.
\\
\par \textbf{Novel Class Discovery without Forgetting} (NCDwF) \cite{2207.10659} is another domain that relaxes some of the assumptions behind NCD.
In NCDwF, $D^l$ and $D^u$ are not available simultaneously.
Instead, during training, we are first given $D^l$ to train the standard supervised task of discriminating known classes.
Then, $D^l$ becomes unavailable and we are given $D^u$ with the goal of discovering the unknown classes.
At inference time, the learned model is evaluated for its performance on instances from a mix known and unknown classes.
This task also poses a greater challenge than NCD as it needs to recognize instances from the full class distribution $\mathcal{Y}^l \cup \mathcal{Y}^u$.
And it is more challenging than GCD as the two training sets $D^l$ and $D^u$ are not available at the same time.
This means that the partitioning of $D^u$ must be learned while avoiding \textit{catastrophic forgetting} on known classes (hence the name).
This domain can be applied if, for example, a model that was previously trained to identify some classes in a dataset that is no longer accessible, and we need to detect new classes while maintaining accuracy on the previously learned categories.

ResTune \cite{liu2022residual} is the first to solve NCDwF.
This article examines three distinct test cases, with NCD and NCDwF among them.
This two-stage method starts with pre-training using the labeled data $D^l$ and a simple cross-entropy loss.
Then, during the training on $D^u$ only, the previously learned representation and classifier are frozen to avoid both forgetting of known classes and overfitting on the unlabeled data.
The partitioning is done by adapting DEC \cite{xie2016unsupervised} to the NCDwF setting.

In \cite{2207.08605}, this problem is referred to as \textit{class-incremental novel
class discovery} (class-iNCD).
Given the NCDwF setting, a two-stage method that seeks to define a classifier capable of predicting in the full label space $\mathcal{Y}^l \cup \mathcal{Y}^u$ is proposed.
Similarly to ResTune \cite{liu2022residual}, an encoder and a classifier are first trained with supervision on the labeled set $D^l$.
Then, during the exploration of the unlabeled set $D^u$, the previously learned classifier is extended with $C^u$ new output neurons.
Additionally, a classification network is added on the shared latent space to partition the unlabeled samples. It is trained with the unsupervised BCE objective of Equation~\eqref{eq:modified_bce} and pseudo labels defined by the RankStats method \cite{autonovel2021}.
The classes predicted by this network are used as targets for the full classification network.

Finally, \cite{2207.10659} introduces the name NCDwF.
To avoid the forgetting, it proposes a method to generate synthetic samples that are representative of each known class and act as a proxy for the no longer available labeled data.
Furthermore, the authors propose a mutual-information based regularizer which improves the partitioning of novel categories, and a Known Class Identifier that helps generalize inference when the test data includes instances from both known and unknown classes.
\\
\par \textbf{Novel Class Discovery in Semantic Segmentation} (NCDSS) is a task defined in \cite{2112.01900} which consists in segmenting images that contain novel classes, given a set of labeled images with known foreground and background classes.
Since the pixels of multiple categories within a single image must be correctly classified, it is more challenging than NCD.
Similarly to NCD, the condition that $\mathcal{Y}^l \cap \mathcal{Y}^u = \emptyset$ is respected, meaning that no image in the unlabeled set contains an object from the known classes.
The framework they propose has three stages: base training, clustering with pseudo labels, and novel fine-tuning.
In the base training stage, the model is trained with labeled base data, which is then used in novel images to filter out salient base pixels and assign base labels.
In the clustering stage, novel images are fed into the model to obtain novel foreground pixels, which are then used for clustering and assigning novel labels.
To address the issue of noisy clustering pseudo labels, an Entropy-based Uncertainty Modeling and Self-training (EUMS) framework is proposed to improve the novel fine-tuning stage by dynamically splitting and reassigning novel data into clean and unclean parts based on entropy ranking.

\section{Tools for Novel Class Discovery}
\label{sec:tools}

Some specific learning paradigms are often found in NCD works.
Namely:
(i) Self-Supervised Learning (SSL) is a popular approach for initializing an encoder,
(ii) Pairwise pseudo labels are used in almost all NCD methods to provide a weak form of supervision for classification neural networks,
and (iii) contrastive learning has been employed by some to construct meaningful and discriminative representations.
In this section, these 3 key paradigms to design NCD methods are presented and discussed.

\subsection{Self-Supervised Learning}

As illustrated in Table~\ref{tab:methods_characteristics}, many methods rely on similarity measures in the latent space to define pairwise relationships between unlabeled instances.
To avoid measuring the similarity after projection through an encoder that was randomly initialized, some methods train for a few epochs with cross-entropy on the labeled samples only.
However, this could result in features that are highly biased towards the labeled data and that poorly represent the unknown classes.
Instead, recent methods have taken advantage of Self-Supervised Learning (SSL) to bootstrap their latent representation.

SSL is a technique that is widely used in computer vision and natural language processing.
The general idea behind SSL methods is to define \textit{pretext tasks} that do not require labels.
A pretext task is a fake problem that can be defined depending on the type of data that is used.
For example, predicting the angle of rotation of an image \cite{gidaris2018unsupervised}, re-coloring \cite{zhang2016colorful} and completing masked words in sentences \cite{devlin2019bert} are common pretext tasks.
Intuitively, SSL allows the model to exploit larger amounts of data by using both labeled and unlabeled data.
The model pre-trained this way will be able to extract more interesting properties, subtle patterns and less common representations of the data, resulting in improved performance compared to solely relying on labeled data.

In the context of Novel Class Discovery, SSL allows the model to learn a robust representation that isn't biased towards the known classes, as all of the data (labeled and unlabeled) is used.
Among SSL methods, RotNet \cite{gidaris2018unsupervised} has been a popular choice in NCD works \cite{autonovel2021, zhong2021neighborhood, zhao2021novel}.
It is a simple and efficient method where the network must predict the rotation angle, from 0, 90, 180 or 270 degrees, applied to an image.
DINO (for \textit{self-\textbf{di}stillation with \textbf{no} labels}) \cite{dino} has also been used in the context of GCD \cite{vaze2022generalized}.
It employs a self-distillation scheme where a student network learns from a teacher given different crops of the same image.
It is a powerful method for vision transformers that produces feature representations where similar objects are close to each other, which is ideal for NCD applications.
Finally, VIME (for \textit{\textbf{v}alue \textbf{i}mputation and \textbf{m}ask \textbf{e}stimation}) \cite{vime} has been used by TabularNCD \cite{tabularncd} to pre-train dense layers in the context of tabular data by reconstructing corrupted samples.
However, as SSL still struggles to be applied to domains such as tabular data, it has only marginally improved performance.
This is partly due to the fact that SSL methods rely heavily on the spatial and semantic structure of image or language data to design pretext tasks.
Thus, only a few works have been proposed to deal with heterogeneous data \cite{vime, bahri2021scarf, ucar2021subtab}.

\subsection{Pseudo labels}
\label{sec:pseudo_label_def}

Pseudo labeling is a technique that provides ``weak'' labels for unlabeled data.
It is particularly useful to exploit large amounts of unlabeled data with models that require a target to be trained.
Apart from NCD, pseudo labels (sometimes called \textit{soft labels}) are found in other domains, such as Semi-Supervised Learning where unlabeled samples that were predicted with high confidence are added to the training data \cite{arazo2020pseudo}. 
In Deep Clustering, they are used to iteratively refine a latent representation by predicting these labels \cite{1604.03628, 1705.07091}.

As expressed in Section~\ref{sec:taxonomy}, most NCD methods define \textit{pairwise} pseudo labels to represent the relationships between pairs of instances in the unlabeled set $D^u$.
In the case of learned-similarity--based NCD methods, they are a way of directly transferring knowledge from the known classes (see Section~\ref{sec:learned_similarity_based}). 
For the rest, pairwise pseudo labels are defined and used in a manner similar to Deep Clustering methods, where they provide supervision for a classification network tasked to partition the unlabeled data\footnote{As this classification network is trained on unlabeled data using these pseudo labels, it is referred to as a ``clustering'' network instead.}.
Instead of directly assigning class labels to instances, the model is only tasked to predict the same label for ``positive'' relations and a different class for ``negative'' relations.
This conversion to a different task is called \textit{problem reduction} \cite{allwein2000reducing}.
It is considered as a less complex problem to solve and to have a lower cost to collect the target.
All pseudo labeling techniques that rely on a similarity measure make the assumption that instances close to each other (usually in the latent space) are likely to belong to the same class.
Pairwise pseudo labels are defined in $\{0, 1\}$ and can be compared for example to the inner product of the prediction through the binary cross-entropy (see Equation~\eqref{eq:modified_bce}).

\begin{figure}[H]
    \centering
    \subfigure[Representation of the data points of the batch.]{\label{fig:pseudo_labels_1}\includegraphics[width=0.21\textwidth]{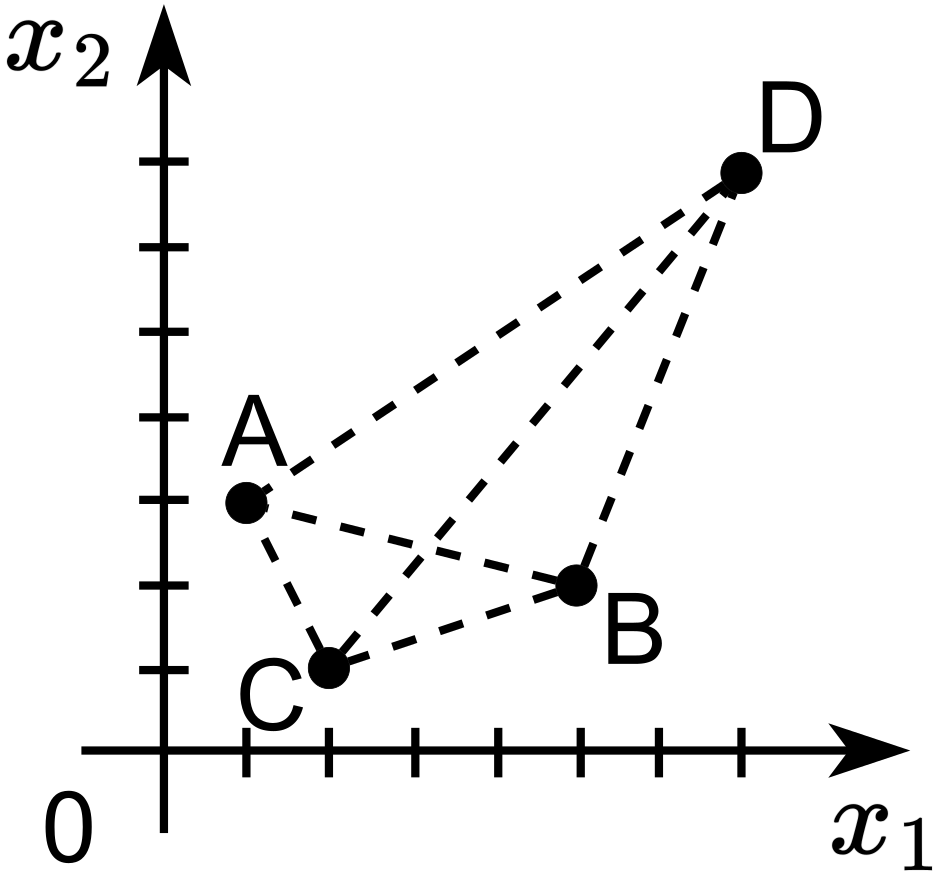}}
    \hspace{0.05\textwidth}
    \subfigure[Pairwise similarity matrix.]{\label{fig:pseudo_labels_2}\includegraphics[width=0.21\textwidth]{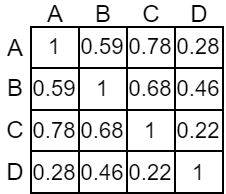}}
    \hspace{0.05\textwidth}
    \subfigure[Pairwise pseudo labels matrix for $\lambda = 0.5$.]{\label{fig:pseudo_labels_3}\includegraphics[width=0.21\textwidth]{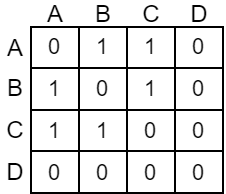}}
    \caption{The pairwise pseudo labels definition process.}
    \label{fig:pseudo_labeling_process}
\end{figure}

To aid the reader in his understanding, Figure~\ref{fig:pseudo_labeling_process} illustrates a simple pseudo labeling process employed by OpenMix \cite{zhong2020openmix} and NCL \cite{zhong2021neighborhood}.
Given a pair $(x_i^u, x_j^u)$ in a batch (Figure~\ref{fig:pseudo_labels_1}), the latent representation $(z_i^u, z_j^u)$ is extracted and their cosine similarity $\delta(z_i^u, z_j^u) = z_i^u \cdot z_j^u / \lVert z_i^u \rVert \lVert z_j^u \rVert$ is computed (Figure~\ref{fig:pseudo_labels_2}).
To use this pairwise similarity matrix as a target for the classification network, it needs to be binarized.
And a solution is to set a threshold $\lambda$ for the minimum similarity score required to consider two instances as belonging to the same class (Figure~\ref{fig:pseudo_labels_3}).
In this case, the pseudo labels are defined as:
\begin{equation}
    \Tilde{y}_{ij} = \mathds{1}[\delta(z_i^u, z_j^u) \geq \lambda]
    \label{eq:pseudo_labels_eq}
\end{equation}
Note that OpenMix sets $\lambda$ to $0.9$ and NCL uses 0.95 arbitrarily, but this is a hyper-parameter that can be optimized.
In the remainder of this section, some of the most commonly used pseudo labeling techniques are introduced.
\\

RankStats (for \textit{ranking statistics}) is a pseudo labeling approach introduced in AutoNovel \cite{autonovel2021}.
Instead of computing a scalar product or a difference between vectors, a pair of instances is considered similar if their features that were ``most activated'' by the encoder are the same.
The authors argue that the most discriminative features of an image should have the highest values after projection.
Thus, RankStats tests whether the $k$ highest values of a pair of embeddings are in the same locations:
\begin{equation}
     \Tilde{y}_{ij} = \mathds{1}[\text{top}_k(z_i^u) = \text{top}_k(z_j^u)]
     \label{eq:rankstats}
\end{equation}
top$_k$ is a function that returns the indices of the $k$ largest values in a vector.
The order of the most activated features is not required to be the same.
It must only contain the same \textit{set} of indices, making RankStats more robust to discrepancies among the most discriminative features.
\\

In \cite{2104.12673}, the Winner-Take-All (WTA) hash \cite{yagnik2011power} is used to compare pairs of instances.
WTA is an embedding method that maps vectors to integer codes.
In more detail, the projection $z_i^u$ of an instance $x_i^u$ is randomly permuted, and the index of the largest elements in its $k$ first values is recorded in $c_i^h$.
This process is repeated $H$ times for each sample $z_i^u$ to form the WTA hash code $c_i = (c_i^1, \dots, c_i^h, \dots, c_i^H)$.
Samples are then compared by applying the same set of permutations and counting the number of indices equal to each other:
\begin{equation}
    \Tilde{y}_{ij} = \mathds{1}[\mathbf{1}^T \cdot (c_i = c_j) \geq \mu]
\end{equation}
with $\mu$ a threshold.
For reference, in \cite{2104.12673}, $H$ is set to the size of the embedding (512), $\mu$ is selected empirically to be 240 and $k=4$.

Intuitively, WTA considers many different orders of features, avoiding the comparison to be dominated by high frequency noise or small local regions that are highly activated.
Replacing the RankStats pseudo labeling method in AutoNovel \cite{autonovel2021} with WTA shows only marginal improvements.
But for the NCD method proposed by the authors in \cite{2104.12673}, WTA consistently outperforms other alternatives, such as RankStats, cosine similarity or nearest neighbour.
\\



Lastly, the quality of the pseudo labels has been explored in some articles.
It is often expressed that they can be noisy and unreliable, and as they have a strong influence on the clustering performance, some works have approached this problem.
OpenMix \cite{zhong2020openmix} mixes labeled and unlabeled samples with MixUp \cite{zhang2018mixup} to generate higher confidence pseudo labels.
DualRS \cite{zhao2021novel} focuses on multiple granularity of image crops to improve reliability.
And \cite{qing2021end} proposes utilizing local structure information in the feature space to construct pairwise pseudo labels, as they are more robust against noise.

\subsection{Contrastive Learning}

Contrastive Learning \cite{hadsell2006dimensionality, chen2020simple} is a self-supervised representation learning technique where the objective is to learn a robust representation.
This is done by pulling together similar samples and pushing apart dissimilar samples.
As labels are not available, a positive pair is usually formed of a sample and its augmented counterpart, while negative pairs are formed with the rest of the data.

Contrastive learning can be easily adapted to take into account labeled samples and to produce even higher quality discriminative representations \cite{khosla2021supervised}.
For these reasons, it is an ideal technique for the task of Novel Class Discovery, and some NCD works have already used contrastive terms.
For instance, NCL \cite{zhong2021neighborhood} adapts the contrastive loss to exploit both the labeled and the unlabeled sets into one holistic framework.
Detailed in Section~\ref{sec:ncl}, their overall loss function is composed of (i) the loss of AutoNovel \cite{autonovel2021} to partition the unlabeled data and (ii) two contrastive terms.
The first is the supervised contrastive loss \cite{khosla2021supervised} applied to the labeled data, and the second is the unsupervised contrastive loss for the unlabeled data.
Their method outperforms all other baselines in the comparison, and they show that the contrastive terms help improve the discrimination of the model.

The Noise-Contrastive Estimation (NCE) \cite{gutmann2010noise}, has been employed by the WTA-based NCD method of \cite{2104.12673}.
It is a parameter estimation method initially designed to be an alternative to the expensive softmax function.
Instead of computing the prediction of the model for every class, only the true class and a few other (called \textit{noisy}) classes have to be estimated.
This principle inspired the supervised contrastive loss \cite{khosla2021supervised}, and it is employed in the NCD method of \cite{2104.12673}.
Given a batch of size $n$ and the projection $z_i$ of an instance $x_i$, \cite{2104.12673} defines the following loss:
\begin{equation}
    \mathcal{L}_{NCE} = - \text{log} \frac{\text{exp}(z_i \cdot \hat{z}_i / \tau)}{\sum_n \mathds{1}[n \neq i] \text{exp}(z_i \cdot z_n / \tau)}
    \label{eq:nce}
\end{equation}
where $\hat{z}_i$ is the augmented counterpart of $z_i$, $\mathds{1}[n \neq i]$ is an indicator function evaluating to 1 \textit{iff} $n \neq i$ and $\tau$ is a temperature parameter.
Note that since the projection $z$ is $\ell_2$-normalized, the cosine similarity can be simplified to the inner product.
In the case of the NCD method of \cite{2104.12673}, this NCE loss is used to maintain a latent representation.
Similarly to NCL \cite{zhong2021neighborhood}, the unlabeled data has positive pairs formed by a sample and its augmented counterpart, while negative pairs are formed with all other samples in the batch.
However, compared to \cite{2104.12673}, NCL reports higher accuracy on the CIFAR-100 and ImageNet datasets.
This could be attributed to the fact that NCL defines additional positive pairs by selecting the most similar pairs in a queue of samples.

OpenCon \cite{sun2022open} is a method proposed for the Generalized Category Discovery problem, where the authors employ class prototypes to separate known and novel classes.
All instances are assigned to their closest prototype, which allows the definition of a set of pseudo-positives $\mathcal{P}(x)$ and pseudo-negatives $\mathcal{N}(x)$ for each instance $x$.
In conventional unsupervised contrastive learning frameworks, only the augmented counterpart of an instance is used to form a positive pair.
In this case, $\mathcal{P}(x)$ can be used to define a larger number of positive pairs.
Given an anchor point $x$, their contrastive loss is defined as:
\begin{equation}
    \mathcal{L}_{OpenCon} = - \frac{1}{| \mathcal{P}(x) |} \sum_{z^+ \in \mathcal{P}(x)} \text{log} \frac{\text{exp}(z \cdot z^+ / \tau)}{\sum_{z^- \in \mathcal{N}(x)} \text{exp}(z \cdot z^- / \tau)}
    \label{eq:opencon}
\end{equation}
where $\tau$ is a temperature parameter and $z$ is the $\ell_2$-normalized projection of $x$.
Two additional terms are optimized during training: the supervised contrastive loss \cite{khosla2021supervised} on the labeled data $D^l$ and the self-supervised contrastive loss \cite{chen2020simple} on the unlabeled data $D^u$.
During training, the class prototypes are defined as moving averages and cluster assignments are updated after each epoch.

\section{Related works}
\label{sec:ncd_related_domains}

\subsection{Unsupervised Clustering}
\label{sec:unsupervisedclustering}

The NCD problem is closely related to unsupervised clustering.
In both domains, the aim is to find a partition of a dataset where no prior knowledge on the unknown classes is available.
Just like in NCD, a common approach is to consider that the close neighborhood of an instance is likely to belong to the same class. In this case, groups where instances are more similar to each other than they are to other groups are created.
The definition of this similarity can vary a lot depending on the purpose of the study or domain-specific assumptions.
The most widely known methods of clustering are usually unsupervised, however we still distinguish them from the less common \textit{semi-supervised} approach (see Section~\ref{sec:semi_supervised}) that leverages a small amount of information to guide the definition of the clusters.

In the completely unsupervised case, many shallow and deep learning based methods have been proposed.
We refer the reader to \cite{xie2016unsupervised} for fundamental work and \cite{deepclustreview} for a more detailed survey.
Some of the main categories of clustering algorithms are:
Centroid-based algorithms create clusters by determining the proximity of data points to a central vector.
Connectivity-based algorithms group data points into clusters using a tree-like structure.
Distribution-based algorithms model the data with a chosen distribution and form clusters based on the likelihood of data points belonging to the same distribution.
Density-based algorithms define clusters as regions of high data density and consider points in sparsely populated areas as outliers.
Finally, Deep Clustering methods aim at jointly conducting dimensionality reduction (or feature transformation) and clustering, which is done independently in other classical works \cite{deepclustreview}.

As Deep Clustering methods learn rich informative representations while separating data into clusters without supervision, their architectures and loss functions are often close to NCD methods where they are even sometimes used as baselines.
They can be easily adapted to the NCD setting, for example by adding a supervised objective trained on the labeled data from $D^l$ to guide the clustering process.
\\
\par \textbf{Discussion.} As expressed in the introduction, fully unsupervised clustering is not a complete solution to the NCD problem.
Multiple and equally valid criteria to partition a dataset can be used, so the definition of what constitutes a good class becomes ambiguous.
This is why the use of a labeled dataset becomes essential to narrow down what constitutes a proper class and guide the clustering process.
Nonetheless, clustering methods are a frequent building block in NCD methods.
An example of this is \textit{Deep Transfer Clustering} \cite{han2019learning}, where the authors extend \textit{Deep Embedded Clustering} \cite{xie2016unsupervised} by guiding its training process with the known classes.
A few works use \textit{k}-means and its variations for label assignment in the feature space of a deep network \cite{vaze2022generalized, wang2020openworld}.
And \cite{wang2021progressive} employs both $k$-means and spectral graph theory to explore the novel classes.

\subsection{Semi-Supervised Learning}
\label{sec:semi_supervised}

Semi-Supervised Learning \cite{ChapelleSemi} is an instance of \textit{weak supervision}, as it uses a limited amount of information in order to carry out its task. It is often reviewed in Novel Class Discovery articles for the similarity of its setup. Four different scenarios can be distinguished in Semi-Supervised Learning: semi-supervised dimensionality reduction \cite{zhang2007semi}, semi-supervised regression \cite{zhou2005semi}, semi-supervised classification \cite{Sugatoconstrainedkmeans, callut2008semi} and semi-supervised clustering \cite{KiriConstrainedKmeans, pckmeans, mehrkanoon2014multiclass}. Only the last two are relevant for our problem, and they are briefly introduced below.

In \textbf{Semi-Supervised Classification}, only a small portion of the dataset is labeled.
This is a setup that can arise when labeling every instance is too costly, but we still wish to leverage the unlabeled data.
Similarly to supervised classification, the goal is to assign instances to one of the classes seen in training, however traditional supervised classification won't take advantage of the unlabeled data.
In this situation, a more accurate model can often be built using semi-supervised learning.
Examples of such models include \textit{constrained $k$-means} and \textit{seeded $k$-means} \cite{Sugatoconstrainedkmeans,LemaireIJCNN2015initialisation}. They are extensions of $k$-means that use a labeled subset to initialize the centroids of the clusters.
It is important to note that the methods in this domain focus on the classification task, where the classes in labeled and unlabeled sets are the same. This is the main difference with the NCD domain, and the reason why semi-supervised learning methods cannot be transferred to our problem.

In the case of \textbf{Semi-Supervised Clustering}, additional information in the form of ``must-link'' and ``cannot-link'' constraints is usually available.
It indicates if pairs of instances must or must not be placed in the same cluster.
Such relations can be derived from class labels.
Examples of semi-supervised clustering algorithms include 
\textit{COP-Kmeans} \cite{KiriConstrainedKmeans}, \textit{PCKmeans} \cite{pckmeans} and \textit{kernel spectral clustering} \cite{mehrkanoon2014multiclass}.
The Novel Class Discovery problem could be reformulated as a Semi-Supervised Clustering problem by defining must-link and cannot-link constraints.
However, the complete set of constraints can only be defined for the labeled data thanks to the ground-truth labels available.
Only cannot-link constraints can be defined between the labeled and unlabeled data (using the hypothesis that $C^l \cap C^u = \emptyset$), and no constraints can be defined for pairs of unlabeled data.
We do not expect this set of constraints to help the clustering process of the unlabeled data.
Furthermore, most Semi-Supervised Learning methods are modified versions of the $k$-means algorithm, and will also suffer when the clusters are not spherical or when the dimension is too large and the euclidean distances becomes inadequate.
%
\\
\par \textbf{Discussion.} Semi-supervised learning methods require either the classes to be known in advance (in the case of partially labeled data) or known constraints on the observations, which is not the case in NCD.
Recent works \cite{chen2020semi, oliver2018realistic} have also shown that the presence of novel class samples in the unlabeled set negatively impacts the performance of such models.
Some articles address this issue \cite{guo2020safe}, but they do not attempt to discover the novel classes.
As such, semi-supervised works are not directly applicable to the Novel Class Discovery problem.

\subsection{Transfer Learning}
\label{sec:Transfer Learning}

Transfer Learning is an other domain often mentioned in NCD articles.
It is a field of machine learning that aims at leveraging knowledge from a source domain or task to solve a different (but related) problem faster or with better generalization.
In computer vision, Transfer Learning is commonly expressed by starting the training from a model that was pre-trained on the ImageNet \cite{imagenet} dataset.
Two scenarios of transfer learning can be distinguished and they are introduced in Table~\ref{tab:TL}.

\begin{table}[H]
    \begin{center}
        \begin{tabular}{|C{25.5mm}|m{50mm}|m{60mm}|}
            \hline
            \multicolumn{1}{|p{22.5mm}|}{Name} & Definition & Example \\
            \hline
            cross-domain transfer learning
            & Also known as \textit{domain adaptation}, a model trained to execute a task on one domain is used to learn the same task on a different (but related) domain.
            & \textit{The knowledge of a classifier trained to recognize positive or negative reviews on the domain of movies can be transferred to the domain of book reviews} \cite{tao2020toward}. \\
            \hline
            cross-task transfer learning
            & The knowledge gained by learning to distinguish some classes is then applied on other classes of the same domain.
            & \textit{A model that was trained to recognize the 5 first digits of the MNIST dataset can be expected to more effectively learn to distinguish the 5 other digits of MNIST} \cite{zhu2011heterogeneous}. \\
            \hline
        \end{tabular}
        \caption{Overview of the scenarios of transfer learning}
        \label{tab:TL}
    \end{center}
\end{table}

With \textbf{cross-domain transfer learning}, a model can be pre-trained on a different but related source dataset.
This is useful when the target dataset has too few instances to obtain good generalization.
In this context, the ``re-usability'' of the source data depends on the overlapping of the features of the source and target domains.
This idea is explored in \cite{yang_zhang_dai_pan_2020}, where the authors distinguish two categories of approaches.
The instance-based approaches attempt to reuse the source domain data after re-sampling or re-weighting and are sensitive to such overlapping.
And feature-representation-based approaches try to find a good representation for both the source and target domain.

In \textbf{cross-task transfer transfer learning}, the label spaces are different.
In this case, methods learn a pair of feature mappings to transform the source and target domain to a common latent space \cite{5694083, 10.5555/2283516.2283652}.
Another approach is to learn a feature mapping to transform data from one domain to another directly \cite{NIPS2008_0060ef47, prettenhofer-stein-2010-cross}.
%
\\
\par \textbf{Discussion.} NCD can be viewed as an unsupervised cross-task transfer learning task, where the knowledge from a classification task on a source dataset is transferred to a clustering task on a target dataset.
The large majority of Transfer Learning articles require the labels of both the source and target domains to be known in advance, which makes the use of such methods impossible in our context of class discovery.
The Constrained Clustering Network (CCN) \cite{hsu2018learning} is an exception in this regard.
It is a method proposed to solve two different transfer learning scenarios, one of which being a cross-task problem where the labels of the target data that must be inferred are not available.
This is essentially the NCD problem, which eventually led to this paper being recognized as one of the earliest NCD works.

\subsection{Open World Learning}
\label{sec:owl}

Rather than being a domain in and of itself, Open World Learning (OWL) \cite{oodsurvey} is a broad term that encompasses all the domains that live under the \textit{open-world} assumption.
Traditional machine learning tasks focus on \textit{closed-world} settings, where the test instances can only be from the distribution that was seen during training.
This is in opposition to the \textit{open-world} setting, where instances can come from outside of the training distribution.
Some of these domains include Anomaly Detection (AD), Novelty Detection (ND), Open Set Recognition (OSR), Out-of-Distribution Detection (OOD Detection) and Outlier Detection (OD).
They are concerned with either or both of semantic shift (when new classes appear) and covariate shift (when the definition of the known classes changes).

To help the reader distinguish these domains, Table~\ref{table:owl_overview} summarizes a few important criteria.
And a general description of each of the 5 domains is provided below.

\begin{table}[htbp]
    \centering
    \begin{tabular}{|l|c|c|c|c|c|c|c|}
        \hline 
        Need to ...                        & NCD$^1$   & GCD$^2$   & AD$^3$    & ND$^4$    & OSR$^5$   & \makecell{OOD$^6$ \\ Detection} & OD$^7$ \\
        \hline
        recognize OOD instances            & \ding{55} & \ding{51} & \ding{51} & \ding{51} & \ding{51} & \ding{51} & \ding{51} \\
        \hline
        have OOD samples during training & \ding{51} & \ding{51} & \ding{51}/\ding{55} & \ding{55} & \ding{55} & \ding{51} & \ding{51} \\
        \hline
        accurately classify known samples  & \ding{55} & \ding{51} & \ding{55} & \ding{55} & \ding{51} & \ding{51} & \ding{55} \\
        \hline
        discover the new classes           & \ding{51} & \ding{51} & \ding{55} & \ding{55} & \ding{55} & \ding{55} & \ding{55} \\
        \hline
    \end{tabular}
    \caption{Overview of the domains in Open-World Learning.}
    \begin{flushleft}
        \footnotesize{$^1$Novel Class Discovery, $^2$Generalized Category Discovery, $^3$Anomaly Detection, $^4$Novelty Detection, $^5$Open Set Recognition, $^6$Out-of-Distribution, $^7$Outlier Detection.}
    \end{flushleft}
    \label{table:owl_overview}
\end{table}

\textbf{Anomaly Detection:}
Given a predefined ``normality'', the goal of AD is to identify abnormal observations.
The abnormality can originate either from a semantic or covariate shift \cite{ruff2021unifying}. 
For example, given a set of pictures of dogs, a model capable of recognizing if a picture is not a dog (i.e. a picture of a cat) falls under semantic shift AD.
In this case, the normality corresponds to all pictures of dogs.
And a model designed to recognize if a given picture of dog is from a breed seen in training falls under covariate shift AD.
We can see that the key to successfully building an AD model is to precisely define the notion of normality.

Two categories of AD settings can be distinguished: either the training set represents the normality, or the the training set is labeled ``normal'' and ``abnormal''.
The first setting is usually preferred, as anomalous data is often found in limited quantities (or even completely unavailable), which makes unsupervised approaches more attractive than supervised ones.
\\
\par \textbf{Novelty Detection:}
From a clean training set with only instances of known classes, the goal of ND is to identify if new test observations come from a novel class or not.
This problem is very close to Anomaly Detection, but it can be differentiated in two ways:
First, this problem is concerned only with semantic shift (i.e. the apparition of new classes).
And second, it does not consider novel samples as ``anomalies'' that must be discarded, but rather as new learning opportunities from events that were not seen during training \cite{markou2003novelty}.
ND stems from the idea that during training, a model cannot have seen all possible classes. Since this idea is very valid in production, traditional classification models can be difficult to apply, and ND models are more convenient.

However, the authors of \cite{oodsurvey} conclude that the goal of ND is only to distinguish novel samples from the training distribution, and not to actually discover the novel classes.
Therefore, most methods assume that the discovery of the new classes in the rejected examples is either the duty of a human or a task that is outside of the scope of their research.
This is a major difference with Novel Class Discovery (NCD), as ultimately, the goal of NCD is to explore the novel samples.
To the best of our knowledge, \cite{shu2018unseen} is an exception.
In this work, an attempt is made by the system to solve this problem while still addressing the other concerns of open-world learning.
\\
\par \textbf{Open Set Recognition:} The idea behind Open Set Recognition (OSR) \cite{Scheirer2013} is that standard neural networks have a tendency to output high confidence predictions even when confronted with instances from classes that were never seen during training.
OSR therefore tries to detect unknown samples additionally to accurately classifying the known classes.
An example of an OSR system would be an application trained to recognize certain faces to allow entry into a building. Such a system must (i) identify known people and (ii) reject the faces from people it has never seen instead of predicting one of the known faces.
%
%
\\
\par \textbf{Out-of-Distribution Detection:}
Similarly to OSR, OOD Detection originates from the idea that machine learning models can predict labels with high confidence for instances of classes they have never seen during training.
OOD Detection methods also aim to (i) accurately classify samples of known classes and (ii) reject samples from outside the known distribution.
Because the definition of ``distribution'' depends on the application, OOD Detection methods cover a large range of methods.
These methods are generally given both In-Distribution (ID) and Out-of-Distribution (OOD) samples during training (see Table~\ref{table:owl_overview}) to narrow down the definition of ID.
Note that OSR and OOD Detection are very close both in setting and goal.
However, they can be differentiated primarily by the fact that OSR methods are tasked with identifying instances that suffer a semantic shift, but originate from the same source dataset, while OOD Detection methods seek to identify semantically different instances that come from a completely different dataset with non-overlapping classes.
\\
\par \textbf{Outlier Detection:}
OD is a task that deviates from the 4 other OWL tasks defined above, as there is no train/test split and all the data is processed together.
The goal is to detect samples that present a significant semantic or covariate deviation from others according to some measure.
Some of the applications of such methods include network intrusion detection \cite{alrawashdeh2016toward}, video surveillance \cite{xiao2015learning} and dataset pre-processing \cite{van2005data}.
Outlier Detection is a well-studied domain with a large number of proposed methods.
Distance-based methods identify points that are far away from all of their neighbors \cite{dang2015distance}, density-based methods select points in sparsely populated regions \cite{breunig2000lof} and clustering-based methods capture samples that did not fall in any of the major clusters \cite{dbscan}.
\\
\par \textbf{Discussion.} The main objective of Open World Learning (OWL) methods is generally to identify instances that come from a different distribution than the known classes in order to reject them and keep a high performance on known classes.
These methods ignore the rejected instances and do not seek to cluster them into novel classes (see Table~\ref{table:owl_overview}).
Because in the \textit{open-world} setting, the data at training or inference time will be a mix of In- and Out-of-Distribution samples, OWL methods are always at least tasked to recognize Out-of-Distribution samples.
This is not a concern in Novel Class Discovery (which does not belong to OWL), as we are given separate datasets during training and only unknown samples at inference.
Instead, NCD could be seen as an extension from OWL works where, after novel samples were detected, we seek to discover the underlying classes.
But as the main focus of these articles is not relevant to the NCD problem, it is difficult to transfer OWL works to NCD.

However, Generalized Category Discovery (GCD, see Section~\ref{sec:gcd}) can be seen as a domain that is halfway between OWL and NCD.
Like in NCD, methods in GCD are given two separate sets during training: a labeled set of known classes and an unlabeled set of unknown classes.
And like in OWL, test samples in GCD can be either from known or unknown classes.
Generalized Category Discovery is very close to OSR and OOD Detection, as it shares their goal of accurately classifying known samples and identifying unknown samples.
It can, however, be distinguished by the fact that semantically shifted samples originate from the same parent distribution (i.e. they are classes from the same dataset), and it seeks to discover the unknown classes.

As many methods in AD/ND/OSR/OOD Detection/OD can be applied to detect instances that are semantically different from the known classes, they could potentially be used for the task of GCD to distinguish if instances come from known or novel classes.
Such methods could be used in a two stage approach, where test samples would first be designated as belonging to known or unknown classes using OWL methods, and then the samples of unknown classes would be clustered with NCD methods.
However, holistic approaches are usually preferred by researchers and works in GCD seem to be following this path \cite{vaze2022generalized, zheng2022openset, Yang_2022_CVPR, fei2022xcon}.



\section{Conclusion and perspectives}

This survey extensively examined the publications in the new field of Novel Class Discovery.
We formally defined the setup and key components of NCD, and proposed a taxonomy that categorizes NCD frameworks based on the way knowledge is transferred between the labeled and unlabeled sets.
We found that two-stage methods were initially popular, but their risk of overfitting on the known classes encouraged defining single-stage methods, which are now widely adopted.
We believe this taxonomy will help guide future research by giving a clear overview of the families of approaches and techniques that have already been explored.
NCD is a newly emerging field that offers a more practical setting compared to fully supervised or unsupervised methods in certain situations.
This has led to the creation of new domains, which we have also analyzed, as researchers have relaxed their assumptions and devised new challenges inspired by NCD.
Additionally, we identified and presented techniques and tools that are commonly used in NCD.
Finally, since this is a new domain that lies at the intersection of several others, it can become challenging to distinguish NCD from other areas of research.
Thus, we also presented the domains most closely related to NCD and highlighted the main differences.
We hope that this last section will help readers unfamiliar with NCD understand what sets it apart from other domains.

Despite the growing body of work in this area, several questions remain unanswered and some perspectives, in our view, are worthy of further study.
As we have seen in this survey, the majority of NCD works are applied only to image data due to specialized architectures and techniques such as data augmentation and self-supervised learning, which rely on the unique structure of images.
They are partly responsible for the success of NCD methods, and since they are not directly applicable to other data types, most works are still limited to image data.
However, it is worth exploring the potential of applying such methods to other data types such as text, tabular, and others.
%
DTC \cite{han2019learning} has shown that deep clustering methods can easily be transferred to the NCD problem, and we expect that more of them could be adapted and offer a new source of inspiration.
Some procedures have been proposed to determine the number of unknown classes automatically with varying degrees of success.
Ideally, NCD methods should not make the assumption that this number is known in advance, but this is most likely not a limiting factor in real-world scenarios.
We also believe that it is crucial to have a unified benchmark and evaluation protocol, since previous works have shown that the split of known/unknown classes has an influence on the difficulty of the NCD problem \cite{li2022a}.
Lastly, the accuracy of pseudo labeling, which widely used in one-stage frameworks, is a decisive factor to the success of these methods. 
There is still room for improvement in this area, for instance, taking labeled data into account, or taking inspiration from graph theory and spectral clustering.

\bibliography{main}
\bibliographystyle{ieeetr}

\end{document}